\documentclass{article}
\usepackage[numbers,sort&compress]{natbib}

\usepackage{arxiv}
\usepackage{hyperref}       
\usepackage{amsmath}   
\usepackage{amssymb}   
\usepackage{esvect}    
\usepackage{graphicx}
\usepackage{caption}
\usepackage{subcaption}
\usepackage{epstopdf} 
\usepackage{algorithm}
\usepackage{algpseudocode}
\usepackage{lettrine}
\usepackage{array}
\title{Learning Superpixel Ensemble and Hierarchy Graphs for Melanoma Detection}

\date{}

\newif\ifuniqueAffiliation

\ifuniqueAffiliation 

\else
\usepackage{authblk}

\author[1]{Asmaa M. Elwer}%
\author[1,2]{Muhammad A. Rushdi\thanks{Corresponding author: \href{mrushdi@eng1.cu.edu.eg}{\texttt{mrushdi@eng1.cu.edu.eg}}}}
\author[3]{%
	Mahmoud H. Annaby}%

\affil[1]{Department of Biomedical Engineering and Systems, Faculty of Engineering, Cairo University, Giza 12613, Egypt}
\affil[2]{School of Information Technology, New Giza University, Giza 12256, Egypt}
\affil[3]{Department of Mathematics, Faculty of Science, Cairo University, Giza 12613, Egypt}
\fi

 \renewcommand{\headeright}{}
 \renewcommand{\undertitle}{}

\begin{document}
	\maketitle
	
	\begin{abstract}
		Graph signal processing (GSP) is becoming a major tool in biomedical signal and image analysis. In most GSP techniques, graph structures and edge weights have been typically set via statistical and computational methods. More recently, graph structure learning methods offered more reliable and flexible data representations. In this work, we introduce a graph learning approach for melanoma detection in dermoscopic images based on two graph-theoretic representations: superpixel ensemble graphs (SEG) and superpixel hierarchy graphs (SHG). For these two types of graphs, superpixel maps of a skin lesion image are respectively generated at multiple levels without and with parent-child constraints among superpixels at adjacent levels, where each level corresponds to a subgraph with a different number of nodes ($20$, $40$, $60$, $80$, or $100$ nodes). Two edge weight assignment techniques are explored: handcrafted Gaussian weights and learned weights based on optimization methods. The graph nodal signals are assigned based on texture, geometric, and color superpixel features. In addition, the effect of graph edge thresholding is investigated by applying different thresholds ($25\%$, $50\%$, and $75\%$) to prune the weakest edges and analyze the impact of pruning on the melanoma detection performance. Experimental evaluation of the proposed method is performed with different classifiers trained and tested on the publicly available ISIC$2017$ dataset. Data augmentation is applied to alleviate class imbalance by adding more melanoma images from the ISIC archive. The results show that learned superpixel ensemble graphs with textural nodal signals give the highest performance reaching an accuracy of $99.00\%$ and an AUC of $99.59\%$.  
	\end{abstract}
	
	\keywords{Superpixel ensemble graph, Superpixel hierarchy graph, Multi-level graph, Graph learning, Graph thresholding, Melanoma, Feature fusion.}

	\section{Introduction}
	
	Melanoma stands as the most life-threatning  kind among skin cancers, marked by the unconstrained growth of melanocytes, which can lead to metastasis \citep{hasan2023skin}. In particular, melanoma is the $17^{th}$ most common cancer in the world and the $3^{rd}$ most frequently diagnosed cancer in the United States \citep{SkinCanc70_online, AIMAtMelanoma}. The global rates of melanoma are on the rise with $331,722$ new cases diagnosed and $58,667$ deaths in $2022$, and predictions expect the diagnosis of $346,838$ new cases and the death of  $61,634$ cases in $2025$ \citep{IRAC, whoCancerTomorrow, SkinCanc70_online, AIMAtMelanoma}.
	In the U.S., it is projected that new melanoma cases will increase by $5.9\%$, with an expected total of $212,200$ diagnoses, including $107,240$ noninvasive cases and $104,960$ invasive cases in $2025$. Additionally, melanoma deaths in the U.S. are anticipated to rise by $1.7\%$, with an estimated $8,430$ fatalities in $2025$ \citep{american_cancer_society}. 
	\par Melanoma is a significant public health concern globally, and its statistical spread highlights the need for effective detection and intervention strategies. Recent research has focused on enhancing detection techniques to ensure timely diagnosis. Traditional methods, such as biopsies and clinical examinations, remain essential diagnostic tools. However, biopsies are less preferred due to their invasive nature while other non-invasive techniques show typically unsatisfactory sensitivity or specificity levels, and this has prompted the exploration of alternative melanoma detection and classification approaches. Notably, the integration of advanced imaging technologies, artificial intelligence (AI), and molecular biomarkers has revolutionized the landscape of melanoma detection.
	
	\par This paper provides a new framework for detecting melanoma skin cancer via learning superpixel graph structures from images of  pigmented skin lesions. 
	
	By modeling images as graphs, where nodes represent pixels or features and edges signify relationships between them, intricate spatial relationships and patterns can be effectively captured. Graph signals can be constructed as well by associating each node with an appropriate signal. Recently, several techniques in graph signal processing have emerged (\citep{bloemheuvel2020graph}, \citep{shuman2013emerging}) with applications in various fields, including transportation network analysis \citep{deri2016new}, signal denoising \citep{onuki2016graph}, and anomaly detection in epileptic seizures \cite {fan2018detecting}. The graph signal models can be especially advantageous in melanoma detection and classification \citep{annaby2021melanoma}, as they capture different aspects of skin lesions and highlight subtle variations in image features that are critical for accurate diagnosis.
	
	For pigmented skin lesion analysis, graphs can capture the topological lesion structure and how subregions of a lesion affect and interact with one another. However, there are no definite rules on the choice of subregion size. Also, subregions at different scales might contribute useful information for melanoma detection and classification. This paper thus introduces multi-level graph signals as novel structures for lesion representation and melanoma detection and classification. Multi-level graphs have already been explored in the literature for representing data with hierarchical structures \citep{Hendrickson1995multilevel, clarke2024multi}. 
	
	While graph signals have been typically constructed using manual approaches based on human expertise,  new methods based on directly learning graph structures from data have recently emerged \citep{xia2021graph,berahmand2025comprehensive}. These methods better capture the latent network structures in real-world data and  have demonstrated significant potential across various domains, such as social network analysis \citep{cai2022graph}, recommendation systems \citep{ijcai2021p630}, and increasingly in medical image analysis \citep{hu2024interpretable}. 
	\par Our work employs several key components to build robust discriminative features for skin lesion representation and skin cancer detection. First of all, conventional image pre-processing operations are applied to skin lesion images. Then, each lesion is subdivided into subregions via superpixel segmentation algorithms \citep{bibid3}. To improve the robustness and discriminative power of the sought lesion representation, the superpixel segmentation can be applied with multiple levels of superpixel sizes. This leads to the construction of superpixel maps at different levels of refinement. Each collection of superpixel maps corresponding to a certain lesion image is organized in a a multi-level ensemble or hierarchical pattern from the most coarse superpixel map to the most refined one. The superpixel ensembles and hierarchies can be modeled by multi-level graph structure, where a graph is constructed fro each level with nodes and edges represented by superpixels and their corresponding pairwise relations, respectively. While the superpixel hierarchies are constrained by parent-child relationships among superpixels of adjacent levels, superpixel ensembles have no such constraints.	
	Our paper introduces few key contributions towards melanoma detection in dermosopic lesion images:
	\begin{itemize}
		
		\item Developing multi-level superpixel ensemble graph (SEG) models where a superpixel map is  independently constructed for each level.
		
		\item Developing superpixel hierarchy graph (SHG) models. Starting from the most refined superpixel map, the superpixel map at any less refined level is created via merging superpixels of the preceding more refined level. 
		
		\item Employing graph structure learning 
		to learn intra-level weighted connectivity patterns for edges within the same graph level.
		
		\item  Extraction of both vertex and spectral domain features from multi-level ensemble and hierarchical graph  representations.
		
		\item Analyzing and visualizing the contributions of the most discriminative features for melanoma detection. Actually, the detection performance was boosted by traditional features as well as features obtained from different graph levels in the multi-level graph representations.   
		
	\end{itemize}
	\par The rest of this paper is organized as follows. Section\ref{RelatedWork} reviews the literature on  automated melanoma detection methods based on conventional, graph-based, and deep learning techniques. Section \ref{sec:DataSet} describes the dataset used in this work. Then, in Section \ref{sec:Multilevel_graph_construction}, the construction of multi-level superpixel graphs from pigmented skin lesions is introduced. Details are given on the structural connectivity of the superpixels and different potential realizations of the multi-level graph signals. After that, Section \ref{sec:Results} presents extensive experimental results while Section \ref{Sec:Discussion} discusses the results and their implications. Finally, Section \ref{Sec:Conclusion} provides conclusions, limitations, and potential directions for future work.
	
	\section{Previous Work}
	\label{RelatedWork}
	
	
	Skin lesion analysis has evolved significantly over the past decade, driven by advances in computational methods. Broadly, the literature can be grouped into four categories: traditional machine learning approaches, deep learning-based models, hybrid techniques that combine both paradigms, and more recently, graph-based learning frameworks \citep{naseri2025diagnosis, jones2022artificial}. Each category reflects a change in the way researchers approach the challenges of skin anomaly detection and classification in medical imaging. The following sections provide an overview of these developments, highlighting their strengths, limitations, and how they are built upon each other.
	\subsection{Conventional Machine Learning Approaches}
	\label{subsec:Traditional Machine Learning Approaches}
	Initial efforts in skin lesion classification relied heavily on handcrafted features and conventional classifiers such as support vector machines (SVM), decision trees, and k-nearest neighbors (KNN). These models used manually designed descriptors—color histograms, texture metrics, and shape moments—to capture visual patterns. While these techniques performed reasonably well in controlled environments, they often struggled with scalability and generalization across diverse datasets. These limitations paved the way for more automated and data-driven approaches.
	Thanh et al. \citep{thanh2020melanoma} introduced the total dermatoscopic score (TDS) based on the clinical ABCD criteria for melanoma diagnosis.  This score was computed from the extracted features of each segmented ROI lesion in  the ISIC$2017$. The achieved accuracy, sensitivity, and specificity were $96.6\%$, $96.1\%$, and $96.8\%$, respectively.  
	Javaid et al. \citep{javaid2021skin} proposed a machine learning approach for skin lesion segmentation and classification in dermoscopic images. For lesion segmentation, contrast stretching and Otsu's thresholding were applied, followed by the extraction of GLCM, HOG, and color features. The feature dimensionality was reduced using principal component analysis (PCA) while the class imbalance was alleviated using the SMOTE method. Random forests were used for classification achieving a $93.89\%$ precision on the ISIC$2016$ dataset.
	\subsection{Deep Learning-Based Methods}
	\label{subsec:Deep Learning-Based Methods}
	The emergence of deep learning, particularly convolutional neural networks (CNNs), marked a turning point in medical image analysis. These models eliminated the need for manual feature extraction by learning hierarchical representations directly from raw pixels. Deep learning methods have demonstrated superior accuracy and robustness, especially when trained on large annotated datasets. Transfer learning and pretrained architectures further expanded their applicability. However, deep models are data-hungry and often lack interpretability, which has led researchers to explore more flexible and transparent alternatives.
	Mandal et al. \citep{mandal2024robust} extracted and fused deep dermoscopic image features using both Xception and Big Transfer (BiT-M) \citep{kolesnikov2020big}. Elementwise multiplication was applied to extract the most important and versatile feature descriptors. A squeeze-and-excitation network was thus trained on descriptors of the ISIC$2017$ dataset to differentiate among three classes: melanoma, nevus, and seborrheic keratosis. The network reached an accuracy of $79.5\%$ and an F1-score of $79.2\%$.
	\subsection {Fusion of Machine Learning and Deep Learning}
	\label{subsec:Hybrid Fusion of Machine Learning and Deep Learning } 
	Hybrid learning methods combine handcrafted features with deep representations, aiming to leverage the strengths of each. Common strategies include feature fusion, ensemble learning, and multi-branch architectures. Hybrid models often improve performance and reduce dependency on large labeled datasets. 
	Salma et al. \citep{salma2022automated} introduced a CAD system for the classification of benign and malignant skin lesions. First of all, hair and artifacts were eliminated via  morphological filtering, and then lesions were segmented using GrabCut segmentation. The ABCD rule was applied for feature extraction, and then various pre-trained CNNs were used to distinguish between malignant and benign lesions. The best performance was attained by a ResNet$50$ network paired with a SVM classifier. Furthermore,  data augmentation boosted the classification performance, leading to a $99.52\%$ AUC, a $99.87\%$ accuracy, a  $98.87\%$ sensitivity, a $98.77\%$ precision, and a $97.83\%$ F1-score.
	Gajera et al. \citep{gajera2023comprehensive} proposed multiple hybrid melanoma detection models that integrate conventional machine learning classifiers with each of eight CNN architectures: AlexNet, VGG-16, VGG-19, Inception v$3$, ResNet$50$, MobileNet, EfficientNet-B0, and DenseNet-121. These models were evaluated on the ISIC$2016$, ISIC$2017$, PH$2$, and HAM$10000$ datasets. An accuracy of $98.33\%$ and F$1$-score of $96.00\%$ were attained using DenseNet-$121$ with MLP on the ISIC$2017$ dataset. 
	Midasala et al. \citep{midasala2024mfeuslnet} introduced MFEUsLNet, a skin cancer classification model combining multi-level feature extraction with unsupervised learning. This method employed bilateral filtering for noise reduction, K-means clustering for lesion segmentation, and extracted RDWT and GLCM features. A recurrent neural network (RNN) classifier was trained on these features to classify dermoscopic lesions. This approach demonstrated strong performance on the ISIC$2020$ dataset, reaching an accuracy of $99.18\%$, and a false-positive rate of $99.25\%$.
	Mahum et al. \citep{mahum2022skin} proposed a hybrid model for skin cancer detection via fusing conventional features (local binary patterns (LBP)) and deep features (based on Inception V3).  The fused features are then used to train an LSTM network, whose performance was tuned using an Adam optimizer and adaptive learning rate adjustment. The model was evaluated on the $1000$-image DermIS dataset, equally divided between benign and malignant cases, achieving an accuracy of $99.4\%$, a precision of $98.7\%$, a recall of $98.66\%$, and an F1-score of $98\%$.
	Soundarya et al. \citep{soundarya2025novel} introduced a hybrid approach for extracting  handcrafted and deep learning features for melanoma detection in dermoscopic images. Specifically, various classifiers were evaluated on combinations of features of grey-level co-occurrence matrices (GLCM), redundant discrete wavelet transform (RDWT), and deep features of  DenseNet121. This yielded an accuracy of $93.46\%$ with an XGBoost classifier, while an ensemble classifier yielded an accuracy of $94.25\%$.
	
	\subsection{Graph-Based Methods}
	\label{subsec:Graph-Based Learning for Medical Imaging} 
	While conventional, deep, or hybrid learning approaches have shown high performance, they still overlook topological inter-connectivity relations and features of pigmented lesion networks. Graph-based methods can account for  these relations, and this makes them particularly well-suited for modeling the complex spatial organization of melanoma lesions. Thus, graph-based modeling represents a key direction for advancing automated melanoma detection. Graph-based methods offer a fresh perspective by modeling images as structured data.  Instead of handling pixels independently,these approaches may represent image clusters as nodes and their interactions as edges. 
	For example, Sadeghi et al. \citep{sadeghi2010graph} proposed a graph-based approach for detecting pigment networks and diagnosing melanoma. Specifically, each color dermoscopic image is converted to a luminance image where lesion edges are detected, and the resulting edge map is used to construct graph patterns in two stages. First, cyclic subgraphs associated with circular or mesh-like patterns of pigment networks are identified. Then, using the spatial relationships between these subgraphs, a secondary graph is constructed, and its density ratio is used to classify the pigmented lesion as melanoma or non-melanoma lesion. A classification accuracy of $92.6\%$ was achieved on 500 test images.
	Clarke et al. \citep{clarke2024multi} analyzed whole slide  images (WSI) of melanoma cases (obtained from the Cahncer Genome Atlas (TCGA) database) and employed multi-level graph representations with histopathological image analysis. In particular, WSIs were converted into graph structures where nodes represented superpixels, and edges represented spatial relationships between them. Features were extracted from the graphs to capture both local and global patterns within the tumor microenvironment. GNNs were used to extract discriminative graph features for lesion classification. The best classifier attained an area under the ROC curve (AUC) of $0.80$. 
	Fu et al. \citep{fu2022graph} introduced a graph-based model that combines intercategory and intermodality networks for diagnosing melanoma, utilizing various types of dermoscopic and clinical images. This model captured the relationships between different types of skin lesions and their associated modalities.  A seven-point checklist dataset was used, along with the ISIC$2017$ and ISIC$2018$ datasets. Image scaling, normalization, and augmentation were performed on the collected images. Then, CNN-based features were independently extracted from the dermoscopic and clinical images of each modality. The interdependencies among various lesion categories and modalities were thus modeled by graph structures where nodes represented different skin lesion categories (or modalities), while edges illustrated the inter-relationships based on the extracted features. This graph-theoretic model was crafted for multilabel classification, allowing for the prediction of multiple skin lesion types simultaneously. The best overall sensitivity, specificity, and accuracy were  $98.21\%$, $96.43\%$ and $97.32\%$, respectively. 
	Filali et al. \citep{filali2021graph} presented a graph-ranking approach for segmenting pigmented skin lesions using macroscopic images. A weighted graph model was constructed for the pigmented lesions, where the graph weights were set via a linear function that combined regional graph features, and the contribution of each feature is controlled by its regional relevance. Then, two ranking operations were applied to each graph to respectively estimate the similarities of the graph regions to cues from the background and foreground. 
	Annaby et al. \citep{annaby2021melanoma} developed a superpixel graph scheme to detect melanoma in dermoscopic images. Instead of relying on individual pixels, the method generated superpixels, which served as graph nodes connected by edges weighted based on the distance between nodal feature descriptors.  Weighted and unweighted graphs were constructed on the superpixel nodes of each lesion image. The connectivity of the graph nodes was defined using an exponential function depending on the degree of similarity among adjacent nodes. For each superpixel, different nodal signals were defined based on  color, texture, or geometry features. Then, features of the vertex domain, spectral graph domain, and whole lesion images were extracted. For the ISIC$2017$ dataset, this approach led to an accuracy of $97.40\%$, a specificity of $95.16\%$ and a sensitivity of $100\%$. 
	Sadeghi et al. \citep{sadeghi2011novel} proposed a graph-based technique for classification and visualization of pigment networks. Edges in the dermoscopic images were firstly detected using a Laplacian-of-Gaussian (LOG) filter. Then, a graph was constructed from each binary edge map, and the graph was used to search for meshes and cyclic structures of a pigmented lesion. Finally, the presence or absence of a pigment network in a certain  skin lesion is determined based on the graph  density. The accuracy of detecting pigment networks in dermoscopic images reached $94.3\%$. 
	Xu et al. \citep{xu2025pixels} present a modular framework for graph-level anomaly detection (GLAD) on dermoscopic images, combining image segmentation, graph construction, and node feature extraction with state-of-the-art Graph Neural Network (GNN) models. Their study systematically evaluates patch-based and superpixel segmentation (SLICO), edge construction via region adjacency graphs (RAG) and k-nearest neighbors, and node features based on color, texture, and shape properties. Virtual nodes are introduced to represent both the lesion and surrounding tissue, enhancing global context. Using the HAM$10000$ dataset, the authors apply five-fold cross-validation and test three GLAD models—SIGNET, CVTGAD, and OCGTL—across unsupervised, weakly supervised, and fully supervised regimes. Results show that color features alone yield strong performance, while adding texture and shape features improves detection depending on supervision level and model architecture. OCGTL achieves $0.805$ AUC-ROC in the unsupervised setting, rising to $0.872$ with weak supervision and $0.914$ under full supervision.
	
	\section{Dataset Description} \label{sec:DataSet}
\begin{figure}[htbp]
		   \centering
		   \begin{tabular}{cc}
			     \subfloat[]{\includegraphics[width=0.22\textwidth]{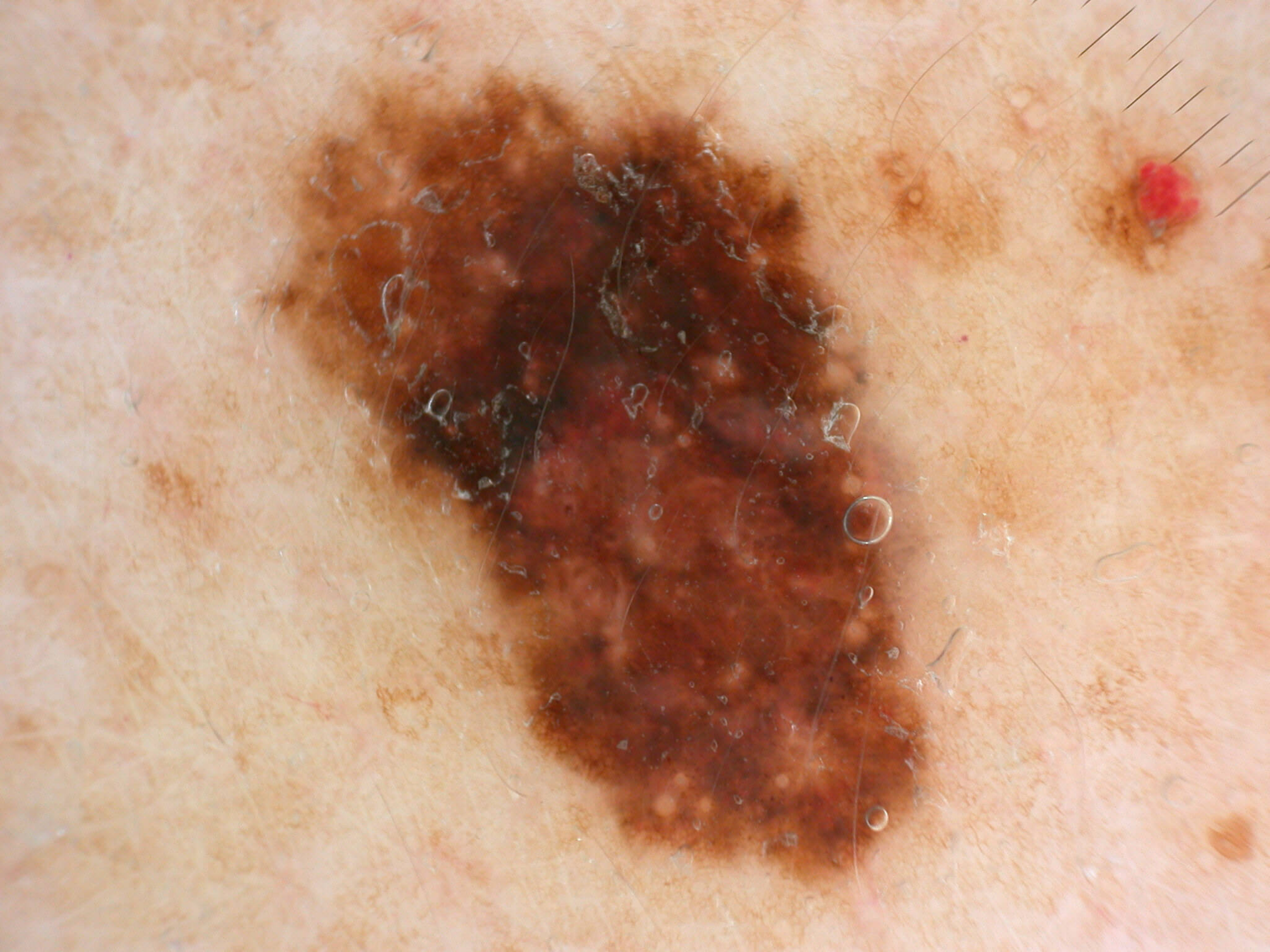}} &
			     \subfloat[]{\includegraphics[width=0.22\textwidth]{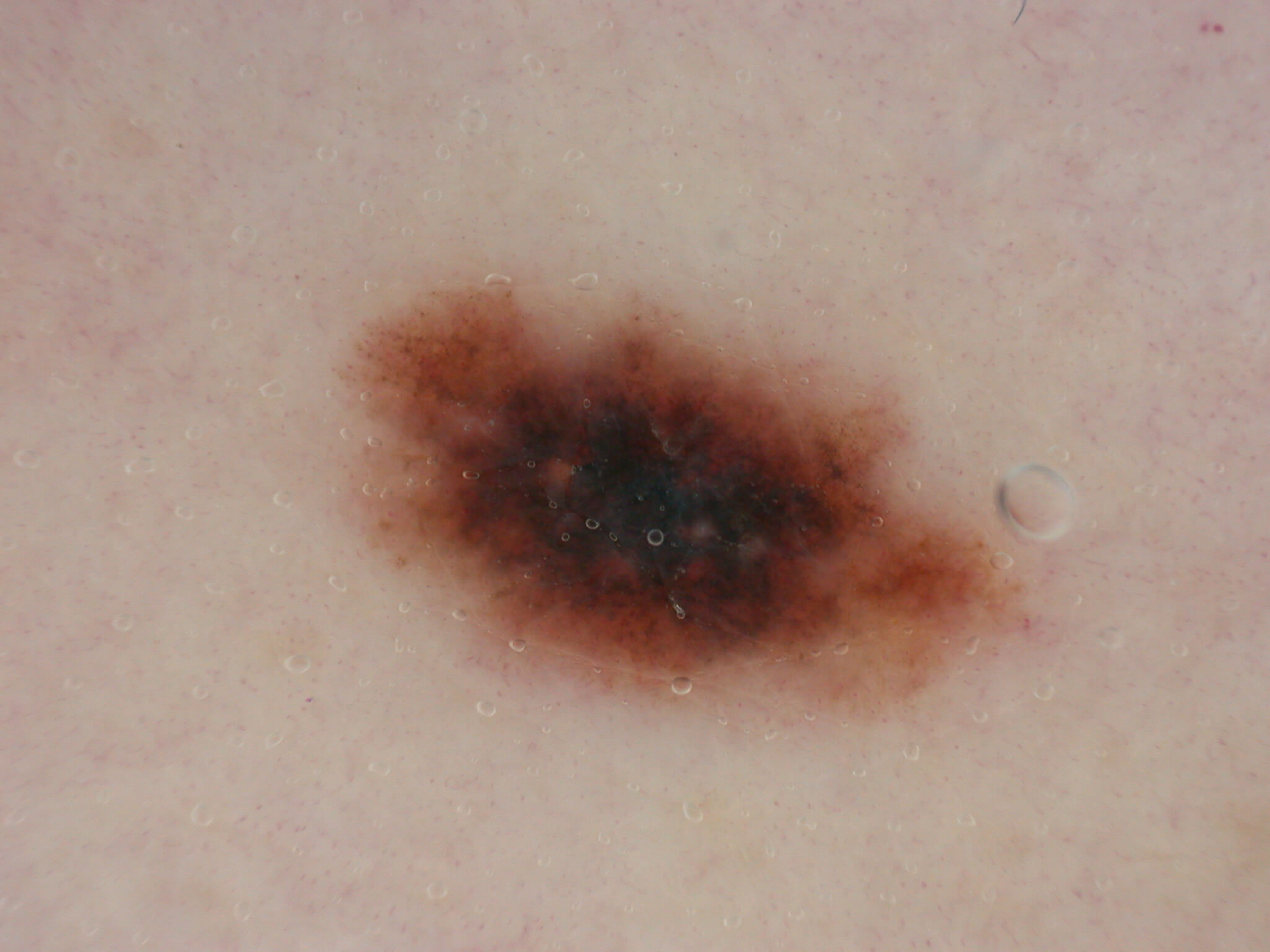}} \\
			     \subfloat[]{\includegraphics[width=0.22\textwidth]{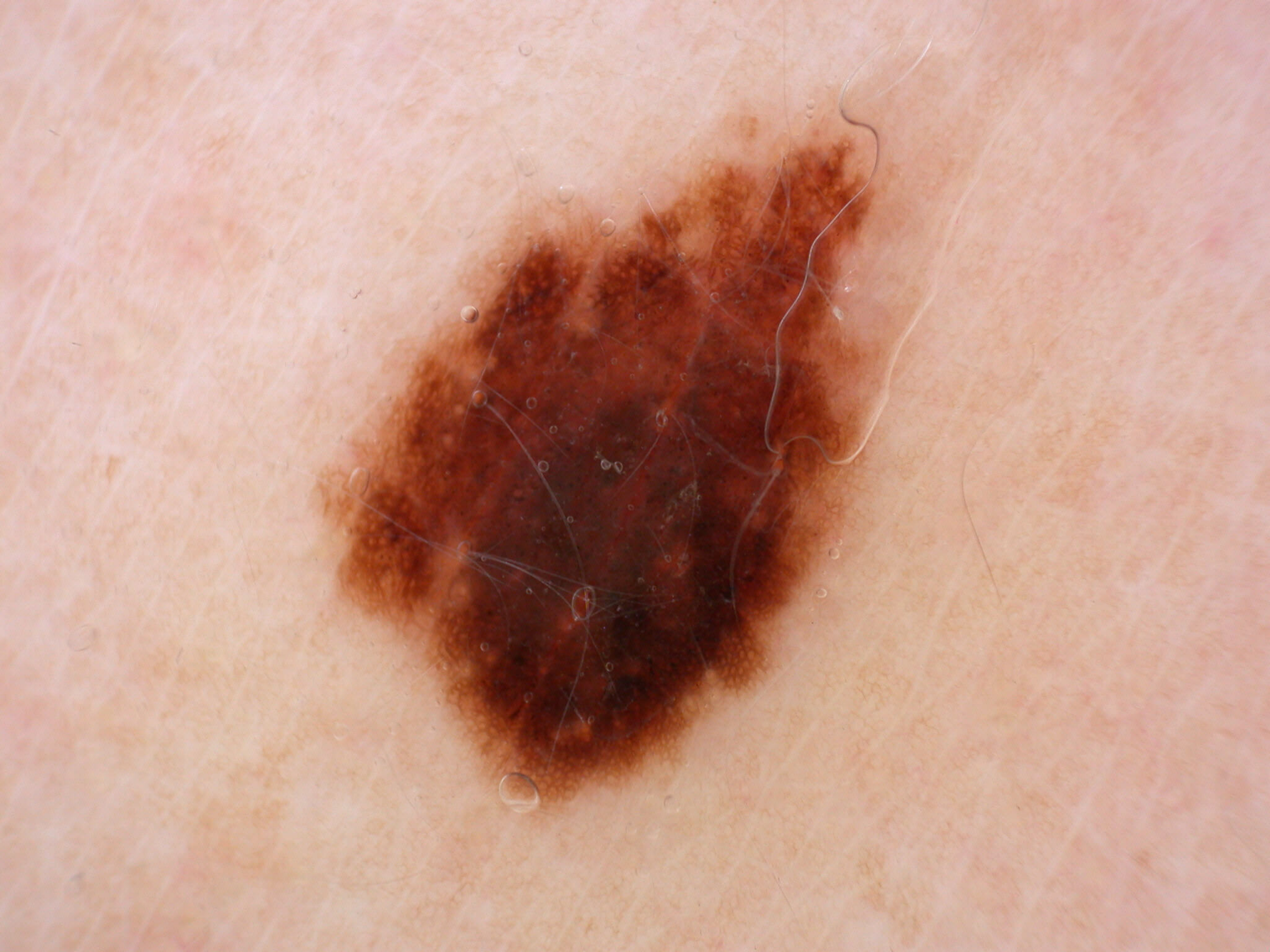}} &
			     \subfloat[]{\includegraphics[width=0.22\textwidth]{S2_Benign.pdf}} \\
			   \end{tabular}
		   \caption{Dermoscopic image samples from the ISIC$2017$ dataset for both melanoma lesions (a, b) and benign lesions (c, d).}
		   \label{fig:onecolumngrid}
		 \end{figure}

	To evaluate the performance of the proposed approach and other recent approaches for melanoma detection, we employed the 2017 dataset of the International Skin Imaging Collaboration (ISIC$2017$) \citep{berseth2017isic}. This dataset has a total of $2600$ dermoscopic skin lesions with $491$ melanoma and $2109$ non-melanoma lesions. Samples of this dataset are shown in Fig. \ref{fig:onecolumngrid}. To alleviate the problem of class imbalance, the dataset is augmented via increasing the images of the ISIC$2017$ dataset with $300$ melanoma images from the ISIC archive \citep{matsunaga2017image}.
	
	\section{multi-level superpixel graphs}
	\label{sec:Multilevel_graph_construction}
	\subsection{Superpixel Graph Construction}
	\label{sec:Superpixel_graph_construction}
	In the following, $G_k=\langle V _k,\varepsilon_k ,W_k \rangle$ is a multi-level family of graphs $k=1,2,3,\dots,N$. Here, $V_k$ is a set of $n_k$ nodes (vertices), and $\varepsilon_k$ is a set of edges and $W_k \in \mathbb{R}^{n_k \times n_k}_{+} $ is the matrix of weights. The multi-level graphs are considered to be simple and undirected. Thus $\varepsilon_K$ is a subset of the set of the unordered pairs $\{\{e_{ij}\}: 1 \leq i,j\leq n_k\}$, $e_{ij}$ is the edge that joins the nodes $v_i$ and $v_j$. Moreover, $e_{ii} \not \in \varepsilon_k$. The vertices $v_i$ are situated at the superpixels of each level of graphs. Each graph level $G_k$ is accompanied with data matrix $X_k = (\vv{X}_1, \dots, \vv{X_{k}})^\mathsf{T} \in \mathbb{R}^{n_k \times m}$. The data matrix $X_k$ depends on the graph level only in size, but it is constructed using color, geometric, and texture properties of the images.
	
	For the completion of the graph construction, a weight matrix $W_k$ must be constructed. There are two ways to construct such a matrix. In \citep{annaby2021melanoma}, the authors used a Gaussian kernel function to construct $W_k$ such that this matrix reflects resumblance between nodes. The other way, which is a major contribution of the present work, is to implement a learning technique to construct $W_k$. Both techniques are outlined in the next two subsections. From now on, we consider a graph $G_=\langle V,\varepsilon,W \rangle$ without the index $k$ for simplicity.
	\subsection{Handcrafted Structured Superpixel Graphs}
	\label{SubSec_Manually structured superpixel graphs}
	Given a graph $G=\langle  V,\varepsilon,W \rangle$, with a set of nodes $V=\{v_1,\dots,v_n\}$ and a set of edges $\varepsilon \subseteq \{\{e_{ij}\}: 1\leq i,j\leq n\}$, we want to associate to $G$ a weight matrix $W \in \mathbb{R}^{n \times n}_{+}$ of non-negative real numbers. Each node $v_i$ is assigned a feature vector $\vv{X_i} \in \mathbb{R}^{m}$. depending on color, geometric, and texture features of the associated superpixels. Thus, a data matrix $X=(\vv{X_1}, \dots, \vv{X_n})^\mathsf{T} \in \mathbb{R}^{n \times m}$ is associated to $G$.
	
\begin{figure}[!t]
	 	  \centering
	 	\begin{tabular}{ccc}
	 		\subfloat[]{\includegraphics[width=0.25\textwidth]{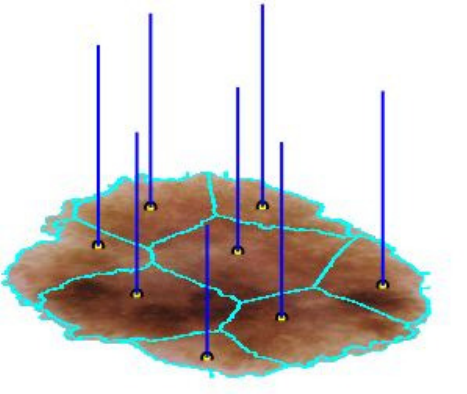}} &
	 		\subfloat[]{\includegraphics[width=0.22\textwidth]{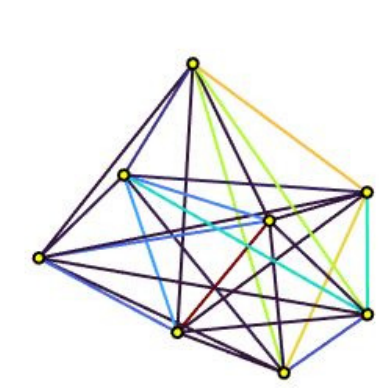}} &
	 		\subfloat[]{\includegraphics[width=0.33\textwidth]{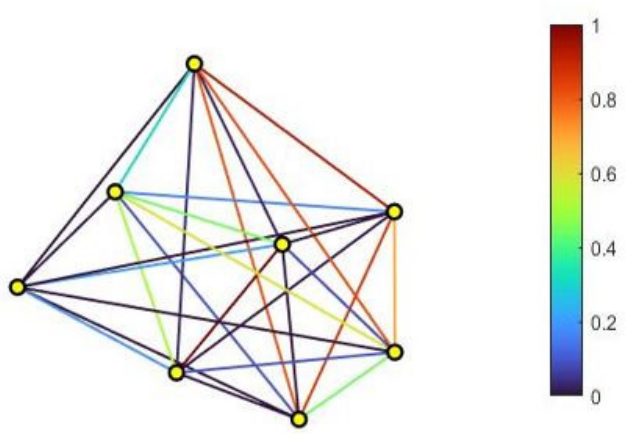}} 
	 	\end{tabular}	
		 	\caption{Superpixel graph construction:  (a) a color graph signal with each bar length representing the signal value for the corresponding superpixel, (b) superpixel graph construction with Gaussian-kernel-based handcrafted weights, and (c) superpixel graph construction with majorization-minimization-based learned weights.}
		 	\label{fig:Two_weight_Methods}
		 \end{figure}

	The first approach to construct $W$ is established manually via a Gaussian kernel, which assigns higher weights to edges between nodes of similar superpixels. Indeed, let $\mu$ and $\sigma^2$ be the mean and variance of the data matrix $X$. The pairwise distance between feature vectors of nodes $v_i$ and $v_j$ can be defined as:
	
	\begin{equation} \label{eq:Eucliden_Distance}
		d_{ij}= \lVert \vv{x_i} -  \vv{x_j} \rVert,  i,j=1, \dots, n, 
	\end{equation}
	where $\lVert.\lVert$ is the Euclidean distance in $\mathbb{R}^m$. The weight matrix $W=(w_{ij})_{1\leq i,j\leq n}$ is defined as \citep{annaby2021melanoma}:
	\begin{equation} \label{eq:Gauss_function}
		w_{ij}=  e ^ { -\left( \frac{( d_{ij}-\mu)^2}{2\sigma^2} \right) }, 1\leq i,j\leq n,
	\end{equation} 
	Figure \ref{fig:Two_weight_Methods} (a, b) show respectively a superpixel color graph signal and the corresponding edge weights set for a Gaussian weighting. The edge color varies from blue (for low edge weights) to red (for high edge weights). 
	
	\subsection{MM-Learned Superpixel Graphs}
	\label{SubSec_Learned superpixel graphs}
	The second approach to construct $W_k$ is established through the learning-based approach of \citep{kalofolias2016learn,sadeghi2010graph}. We let $G,n,m,X$ have the same meaning of Subsection \ref{SubSec_Manually structured superpixel graphs}. In addition, we denote by $D=(d_{ij})_{1 \leq i,j \leq 2}$ to the distance matrix between nodes ( Eq. \ref{eq:Eucliden_Distance}). $\mathbf{1}=(1 1 \dots 1)^\mathsf{T} \in\mathcal{R}^n$, $W\bold{1} \in \mathcal{R}^m$ is the vector of edges degree.
	
	kalofolias \citep{kalofolias2016learn} suggested the following learning approach to construct $W$
	
	\begin{equation}\label{eq:min_Fn}
		\begin{aligned}
			\min_{W \in \mathbb{R}^{n \times n}} \quad & \lVert WD \rVert_{1} - \delta \mathbf{1}^\mathsf{T} \log(W \mathbf{1}) + \frac{\gamma}{2}\lVert W \rVert_{F}^{2} \\
			\text{s.t.} \quad & W = W^\mathsf{T}, \text{diag}(W) = 0.
		\end{aligned}
	\end{equation}  
	The first term reflects smoothness of the constructed graph  signal, and the second term assumes that every node must have a positive degree, while the last term penalizes matrices with high weights. Here, $\lVert\cdot \rVert_1$, $\lVert\cdot \rVert_F$ are the $L_1$ and Frobenius norms respectively \citep{horn2012matrix}. The numbers $\delta, \gamma$ are regularization parameters. 
	
	Among several techniques to solve Eq. (\ref{eq:min_Fn}), Fatima et al. \citep{fatima2022learning} introduced a fast algorithm based on the majorization-minimization (MM) optimization technique \citep{lange2016mm}. Here, we describe the MM algorithm briefly. Since $W, D$ are symmetric, then the objective function is equivalent to 
	\begin{equation}\label{eq:min_Fn_1}
		f(\vv{w})= 2 \vv{w}^\mathsf{T} \vv{d}  - \delta 1 ^\mathsf{T} \log(T\vv{w}) + \gamma \lVert \vv{w} \rVert^{2},
	\end{equation}  
	where $\vec{w}, \vec{d} \in \mathbb{R}^r$, $r=\frac{n(n-1)}{2}$, are vectors formed from the upper (or lower) off-diagonal elements of $W$ and $D$, respectively. Here $T=(\vec{t}_1 \dots \vec{t}_n)^\mathsf{T} \in \mathbb{R}^{n\times r}_{+}$ is a binary matrix for which $T\vec{w}=W1$. Thus problem (\ref{eq:min_Fn}) is equivalent to 
	\begin{equation}\label{eq:min_Fn_2}
		\begin{aligned}
			\min_{\vec{w} \in \mathbb{R}^{r}} \quad &  2 \vec{w}^\mathsf{T} \vec{d} - \delta \sum_{i=1}^{n} \log(\vec{t}_{i}^{\mathsf{T}} \vec{w}) +  \gamma \lVert \vec{w} \rVert ^{2},
		\end{aligned}
	\end{equation}  
	
	Using the convexity of $-\log x$, Fatima et al. \citep{fatima2022learning} gave an algorithm to construct a sequence of vectors $\vec{w_0},\vec{w_1}, \dots$, for which $f(\vec{w}_k$ is a non-increasing sequence that converges to the minimum of $f(vv{w})$, $\vv{w}\in \mathbb{R}^r$. They have constructed an appropriate surrogate function $g(\vv{w}|\vv{w}_j)$ and proved that if $\vv{w}=(w_1, \vv, w_r)$, $\vv{d}=(d_1, \dots ,d_r)$, then the $(j+1)$st iteration of $w_i$ 
	
	\begin{equation} \label{eq:Weight_Solution_min_fn}
		w_{i}^{\left(j+1\right)}=\frac{-2d_{i}+\sqrt{4d_{i}^{2}+8\gamma c_{i}}}{4\gamma} 
	\end{equation}
	where
	\begin{equation} \label{eq:Weight_Solution_min_fn_1}
		c_{i} = \delta \sum_{i=1}^{n} \frac{t_{ji} w_i^{(j)}}{t_j^\mathsf{T} w^{(j)}}
	\end{equation}
	This algorithm is listed in Algorithm \ref{Alg:AlgorithmPseudoCode_W}. Figure \ref{fig:Two_weight_Methods}(b), shows graph construction based on the MM learning technique for the color graph signal in Figure \ref{fig:Two_weight_Methods}(a).
	\begin{algorithm}[H]
		\caption{Learning Single Superpixel Graph Edge Weights}\label{Alg:AlgorithmPseudoCode_W}
		\begin{algorithmic}[1]
			\State \textbf{Input:} $X$, $\varepsilon$, $\delta$, $\gamma$
			\State \textbf{Initialize:} $\mathbf{w}^{(0)} \gets \mathbf{1}$ \Comment{Set all edge weights to 1}
			\State $k \gets 0$
			\Repeat
			\For{$i = 1$ to $n$}
			\For{$j = i+1$ to $n$}
			\State Compute dissimilarity: $d_{ij} \gets \|X_i - X_j\|$
			\EndFor
			\EndFor
			\State Compute surrogate coefficients $\{c_j\}_{j=1}^m$ based on current $\mathbf{w}^{(k)} $ and superpixel signal $X_i$.
			\For{$e = 1$ to $m$}
			\State Update edge weight: $w^{(k+1)}_e \gets \frac{-2d_e + \sqrt{4d_e^2 + 8\gamma c_e}}{4\gamma}$
			\EndFor
			\State Evaluate objective: $f(\mathbf{w}^{(k)})$ and $f(\mathbf{w}^{(k+1)})$
			\State $k \gets k + 1$
			\Until{$\left| \frac{f(\mathbf{w}^{(k-1)}) - f(\mathbf{w}^{(k)})}{f(\mathbf{w}^{(k-1)})} \right| \leq \epsilon$}
			\State \textbf{Output:} Optimized edge weights $\mathbf{w}^{(k)}$
		\end{algorithmic}
	\end{algorithm}

	\subsection{ Multi-Level Superpixel Graph Representation}
	\label{sec:Multi_Level_hierarchical_superpixel_graph}
\begin{figure*}[]
		 		 	  \centering
		 	\begin{tabular}{c}
		 		\subfloat[]{\includegraphics[width=\textwidth]{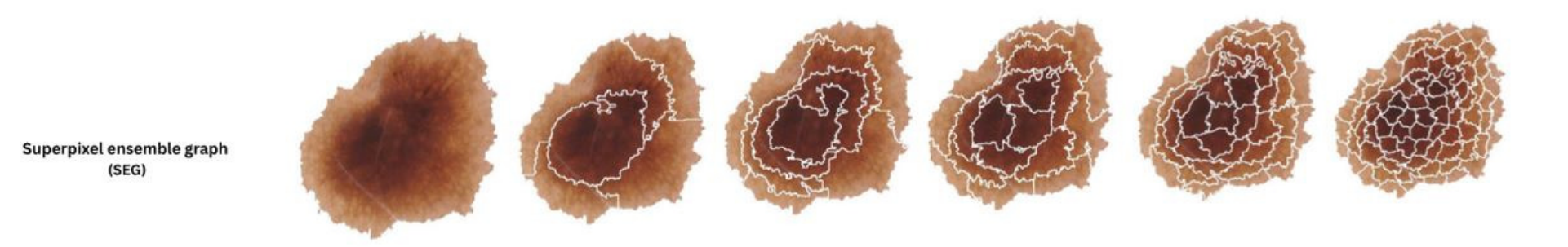}} \\
		 		\subfloat[]{\includegraphics[width=\textwidth]{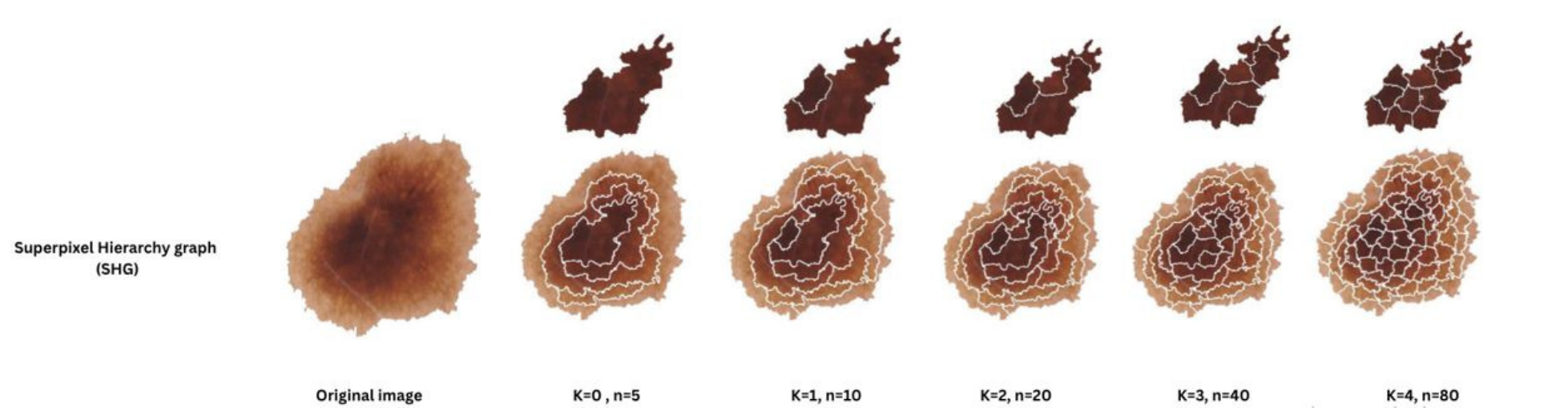}} 
		 	\end{tabular}
		 	
		       \caption{A hierarchical superpixel image structure with $M=80$ superpixels at the most refined level $K=4$. (a) Superpixel ensemble graph (SEG) where superpixel maps are independently constructed. (b) Superpixel hierarchy graph (SHG) where every superpixel map results from the immediately more refined one by merging every pair of superpixels.}
		 	\label{fig:Two_SPLevels}
		 \end{figure*}

	As mentioned above, each dermoscopic image is transformed into superpixel maps with different levels (i.e. numbers of superpixels). Thus, the input image is transformed into a collection of  superpixel maps with $M/2^k$, $k=1,2, \dots, N$ superpixels. Here, $M$ is an initialized value which is taken here to be $100$. We used two techniques to establish superpixel ensemble graphs (SEG) and superpixel hierarchy graphs (SHG). For the superpixel ensemble graphs, we apply the techniques indicated in Subsections \ref{SubSec_Manually structured superpixel graphs} \ref{SubSec_Learned superpixel graphs} independently to produce the superpixel maps $I_1, \dots ,I_N$ of a sample image $I$, i.e. there is no relationship between any two superpixel maps. See Figure \ref{fig:Two_SPLevels}(a).
	
	For the superpixel hierarchy graphs, each superpixel map $I_l+1$ is formed from the map $I_l$ by merging closest pairs of superpixels as shown in Figure \ref{fig:Two_SPLevels}(b) \citep{sharma2005ciede2000}. A subregion of the input image is shown on top of each superpixel map to clarify the subdivision structure. It is worthwhile to mention that the hierarchy of the multi-level structure plays a critical role in the detection system, as it leads to aggregated feature vectors, which improve the classification results, as indicated in Subsection \ref{sec:Learning_Algorithm_Sec}
	
	 \begin{figure*}[!t]
		 	\centering
		 	\includegraphics[width=\linewidth]{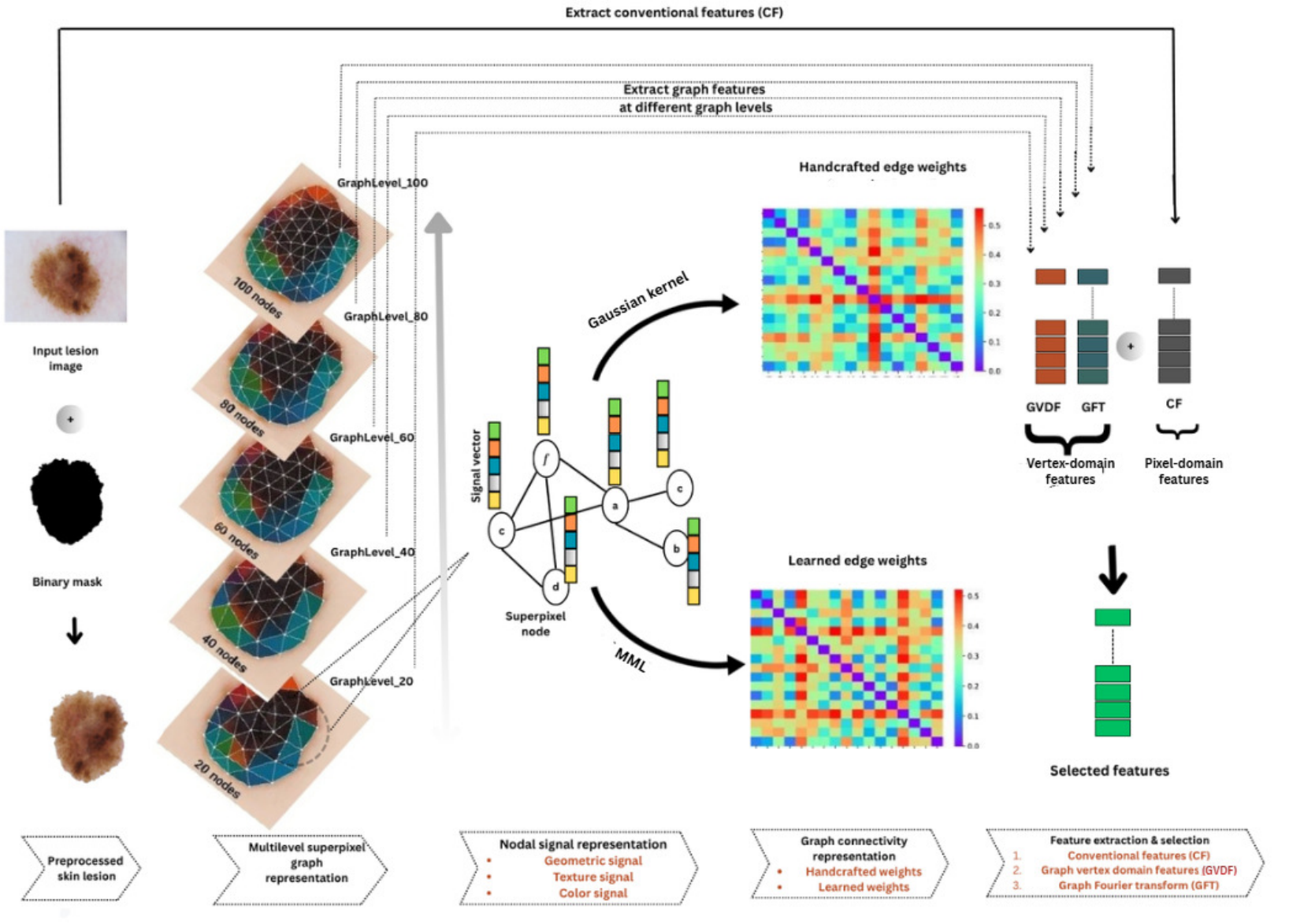} 
		 	\caption{Block diagram of the proposed melanoma detection system.}\label{fig:High_Level_Design}
		 \end{figure*}
	\begin{table*}[h!]
		\centering
		\caption{Melanoma detection results based on single-level graphs and \textbf{superpixel ensemble graphs (SEG)} with \textbf{handcrafted weights} and  \textbf{geometric signals}. For each graph type, the best results are shown in bold. The best overall results are underlined.}\label{tab:Performance_Full_gauss_Geometric}
		\centering
		\footnotesize
		\begin{tabular}{>{\centering\arraybackslash}m{2cm} >{\centering\arraybackslash}m{0.4cm}>{\centering\arraybackslash}m{0.85cm}>{\centering\arraybackslash}m{0.85cm}>{\centering\arraybackslash}m{0.85cm}>{\centering\arraybackslash}m{0.85cm}>{\centering\arraybackslash}m{0.85cm}>{\centering\arraybackslash}m{0.85cm}>{\centering\arraybackslash}m{0.85cm}>{\centering\arraybackslash}m{0.85cm}}
			\hline
			Graph level&	&	SVM   & KNN & QDA & RF & GB & AB  & MLP\\
			\hline

			Single graph&AC& 95.92&	96.21	&87.17	&90.38	&\textbf{97.08}	&95.63&	89.50	\\
			(n=20)&Spec&93.77	&91.73	&98.36	&73.77	&\textbf{98.72}&	98.42	&91.75	\\
			&Sens&91.82& \textbf{97.56} &	81.36	&78.28	&96.83	&92.31	&87.78	\\
			&AUC&  92.33  &  \textbf{97.89}   & 95.67 &   93.29   &89.87&  98.35 &  97.27 \\
			\hline 
			
			Single graph&AC& \textbf{98.04} &	95.92&	87.46&	90.67&	96.50&	96.50	&88.92	\\
			(n=40)&Spec&98.11	&90.30	& 94.85&	\textbf{99.18}&	98.36&	99.18&	98.36	\\
			&Sens&97.74	& \underline{\textbf{98.53}}&81.82	& 98.18	&98.19&	92.31&	87.33	\\
			& AUC & 92.83&  \textbf{ 98.17}&   95.13&   93.68&   92.65& 98.09&  97.64\\
			
			\hline 
			
			Single graph&AC& 97.13	&\textbf{97.38}	&85.13&	90.67&	97.08&	94.46&	90.38	\\
			(n=60)&Spec&98.36	&93.85&	93.04&	99.18&	\underline{\textbf{100.00}}&	98.68&	98.36	\\
			&Sens&95.93	&\textbf{97.31}&	77.38&	89.92&	97.29&	90.95&	86.88	\\
			& AUC &    93.22&   96.49&  97.07&  \textbf{97.88}&   92.24&  97.71&   97.80\\
			\hline 
			
			Single graph&AC& 88.63&	97.38&	86.88	&90.67&	\textbf{98.00}&	93.88&	91.55	\\
			(n=80)&Spec&91.18	&93.13	&97.54	&91.73	&97.38	&98.27	&\textbf{98.36}\\	
			&Sens&95.02&\textbf{97.35}&	83.26&	85.97	&97.29	&90.50	&94.69	\\
			& AUC &    87.76&  97.76&  97.25&   93.86& 88.73& 97.48&  \textbf{97.82}\\
			\hline 
			
			Single graph&AC& 97.08&	96.21	&89.50	&89.80&	\textbf{97.96}	&95.34	&90.96	\\
			(n=100)&Spec&94.26&	90.37	&95.90&	95.14&	96.97&	97.15&\textbf{97.54}	\\
			&Sens&96.82&	92.12	&85.52	&85.97&	\textbf{97.29}	&90.95&	90.05	\\
			&AUC& 
			97.66&  94.26&   94.94&   95.92&  \textbf{97.76}&   96.40&   97.14\\
			\hline
			
			Ensemble graph&AC&  97.13&	96.79	&94.12	& \underline{\textbf{98.10}}	&96.83	&93.00	&94.46	\\
			(fused n)&Spec&96.72&	92.37	&93.39&	 \textbf{98.94}	&95.85	&97.38&	98.36	\\
			&Sens&97.24	& \textbf{97.37}	&89.42&	97.14&	97.19	&95.02&	92.31	\\
			&AUC&
			\underline{\textbf{98.86}}&  88.76&  91.11  &   92.48&  92.65&   91.24&  91.41\\
			\hline 
			
		\end{tabular}
	\end{table*}
	
	
	\begin{table*}[h!]
		\centering
		\caption{Melanoma detection results based on single-level graphs and \textbf{superpixel ensemble graphs (SEG)} with \textbf{handcrafted weights} and  \textbf{texture signals}. For each graph type, the best results are shown in bold. The best overall results are underlined.}\label{tab:Performance_Full_gauss_Texture}
		\centering
		\footnotesize
		\begin{tabular}{>{\centering\arraybackslash}m{2cm} >{\centering\arraybackslash}m{0.4cm}>{\centering\arraybackslash}m{0.85cm}>{\centering\arraybackslash}m{0.85cm}>{\centering\arraybackslash}m{0.85cm}>{\centering\arraybackslash}m{0.85cm}>{\centering\arraybackslash}m{0.85cm}>{\centering\arraybackslash}m{0.85cm}>{\centering\arraybackslash}m{0.85cm}>{\centering\arraybackslash}m{0.85cm}}
			\hline
			Graph level&	&	SVM   & KNN & QDA & RF & GB & AB  & MLP\\
			\hline
			
			Single graph&	AC&96.00	&97.67&	79.59	&90.38&	\textbf{98.05}&	95.63	&90.38	\\
			(n=20)&Spec&73.77&	94.53	&96.72	&73.77&	98.36	& \underline{\textbf{100.00}}&	99.18	\\
			&Sens&94.12	&\underline{\textbf{99.03}}&	72.27&	79.64&	97.74	&91.86&	90.05	\\
			& AUC& 92.92 & 95.34 & \textbf{97.22}  & 93.31 & 89.89 & 94.05 & 92.73 \\ 
			
			\hline
			
			Single graph&	AC&\textbf{97.54}	&97.19	&87.46	&90.67&	93.38&	97.08&	88.92\\	
			(n=40)&Spec&96.72	&93.08	&99.18	&99.18&	99.18& \underline{\textbf{100.00}}&	98.36\\	
			&Sens&93.67	&\textbf{98.85}&	83.71&	97.29	&97.29&	93.67&	87.33\\	
			& AUC& 94.40 & \textbf{98.12} & 91.16 & 93.59 & 96.61  & 97.52 & 97.93 \\ 
			
			\hline
			
			Single graph&	AC&\textbf{97.67}	&96.50	&93.00	&90.38	&97.38&	95.34	&89.80	\\
			(n=60)&Spec&95.90	&93.85	&97.54	&\textbf{99.27}	&99.18	&92.01	&98.36	\\
			&Sens&96.36	&\textbf{98.53}	&89.14&	83.71	&96.38	&91.40&	86.88	\\
			& AUC& 92.60 & \textbf{98.06} & 96.63 & 93.36 & 90.63 & 97.59 & 91.68 \\ 
			
			\hline
			
			Single graph&	AC&88.34&	\textbf{96.92}&	88.63	&90.67&	96.26&	95.63	&90.96	\\
			(n=80)&Spec&97.54	&89.71	&96.72&	\textbf{99.18}&	99.01&	\textbf{99.18}	&98.36	\\
			&Sens&96.38&\textbf{98.63}&	84.16&	84.16	&96.83	&91.86	&90.05	\\
			& AUC& \textbf{98.26} & 95.67 & 94.02 & 97.85 & 98.31 & 89.56 & 98.12 \\ 
			
			\hline
			
			Single graph&	AC&\textbf{97.83}	&95.63	&89.80	&90.09	&97.38&	95.63	&92.42	\\
			(n=100)&Spec&\underline{\textbf{100.00}}&	89.05&	97.54&	99.18&	94.31&	98.67&	97.98	\\
			&Sens&96.82	&\textbf{98.15}	&85.52	&81.00&	96.38	&92.50&	88.69	\\
			& AUC& 92.89 & 97.26 & 97.37 & 98.14 & 91.35 & \textbf{98.39 }& 98.08 \\ 
			
			\hline
			
			Ensemble graph&	AC&\underline{\textbf{98.41}}	&96.79	&85.29	&98.27	&98.25	&94.46	&93.59	\\
			(fused n)&Spec&98.36	&93.02&	92.56	&\underline{\textbf{100.00}}	&90.37&	98.67&	97.54	\\
			&Sens&97.43	&97.55		&81.14	& \textbf{98.57}	&97.29	&96.83	&94.57	\\
			& AUC& 90.95 & 90.91 & 98.64 & 91.37 & 91.32 & \underline{\textbf{98.97}} & 98.93 \\
			
			\hline 
			
		\end{tabular}
	\end{table*}
	
	\begin{table*}[h!]
		\centering
		\caption{Melanoma detection results based on single-level graphs and \textbf{superpixel ensemble graphs (SEG)} with \textbf{handcrafted weights} and  \textbf{color signals}. For each graph type, the best results are shown in bold. The best overall results are underlined.}\label{tab:Performance_Full_gauss_Color}
		\centering
		\footnotesize
		\begin{tabular}{>{\centering\arraybackslash}m{2cm} >{\centering\arraybackslash}m{0.4cm}>{\centering\arraybackslash}m{0.85cm}>{\centering\arraybackslash}m{0.85cm}>{\centering\arraybackslash}m{0.85cm}>{\centering\arraybackslash}m{0.85cm}>{\centering\arraybackslash}m{0.85cm}>{\centering\arraybackslash}m{0.85cm}>{\centering\arraybackslash}m{0.85cm}>{\centering\arraybackslash}m{0.85cm}}
			\hline
			Graph level&	&	SVM   & KNN & QDA & RF & GB & AB  & MLP\\
			\hline
			
			Single graph&AC	&96.79&	96.79&81.34&90.96&	97.96	&\textbf{97.37}	&90.38	\\
			(n=20)&Spec&86.77	&91.73&95.52&75.41& 97.01	&98.67	&\textbf{99.18}	\\
			&Sens&95.02&\textbf{98.01}	&84.55	&79.19&	97.29	&95.93&	86.43	\\
			&AUC & 93.23 & 89.71 & 95.33 & 93.28 & 91.33 & \textbf{98.51} & 91.11 \\ 
			
			\hline
			
			Single graph&AC	&96.41	&97.08	&88.34&	90.96&	\textbf{97.38}	&95.92	&93.59	\\
			(n=40)&Spec&\underline{\textbf{100.00}}&	93.75	&91.80&	99.18&96.27	&99.18&	99.18	\\
			&Sens&96.82&	\textbf{98.54}	&89.14	&98.11	&98.09&	90.05	&90.95\\
			& AUC& 92.36 & 97.57 &88.17  & 94.44 & 93.75 & 91.77 & \textbf{97.77} \\ 
			
			\hline
			
			Single graph&AC	&\textbf{98.83}	&95.92	&88.63&	90.38	&97.38&	95.92&	90.38\\	
			(n=60)&Spec&97.54	&89.71	&98.36&	99.18	&99.18	&\underline{\textbf{100.00}}	&	97.54	\\
			&Sens&96.38&	\textbf{98.60}	&83.26&	87.33	&97.74&	89.59	&88.24	\\
			& AUC& 91.44 & \textbf{98.14} & 97.42 & 94.24 & 89.85 & 97.66 & 90.34 \\ 
			
			\hline
			
			Single graph&AC	&88.63&97.38&	94.75&	90.67&	\textbf{98.25}	&95.63&	88.63\\	
			(n=80)&Spec&98.78&	94.19&	97.56 &	\textbf{99.18}&	97.56	&96.34	&97.34	\\
			&Sens&96.83&	\textbf{98.15}&	92.31&	83.26&	97.74&	90.50	&86.88\\	
			& AUC& 92.70 & \textbf{98.43} & 95.20 & 95.40 & 90.20 & 96.89 & 97.83 \\ 
			
			\hline
			
			Single graph&AC	&96.21	&96.21	&88.63	&90.96&\textbf{	98.25}	&95.92	&90.09\\	
			(n=100)&Spec&98.36	&90.37	&91.80&	99.18	&99.18	&\underline{\textbf{100.00}}&97.54	\\
			&Sens&96.12&	\textbf{98.59}&	86.88	&82.35&	97.29&	93.21&	88.69\\	
			& AUC& \underline{\textbf{99.26}} & 97.96 & 95.57 & 98.24 & 92.26 & 97.90 & 97.96 \\ 
			\hline
			
			Ensemble graph&AC	&98.01&	96.50	&93.77	&\underline{\textbf{99.01}}&	98.83&	95.63&	94.46\\	
			(fused n)&Spec&98.36	&92.31&	98.36	&\underline{\textbf{100.0}}&99.18&	99.18&	96.72	\\
			&Sens&97.73	&98.62&	90.12	&\underline{\textbf{99.13}}&	98.19&	95.93	&93.21	\\
			&AUC & \textbf{99.10 }& 96.19 & 98.10 & 98.85 & 92.75 & 91.94 & 93.09 \\ 
			\hline 
			
		\end{tabular}
	\end{table*}

	\begin{table*}[h!]
		\centering
		\caption{Melanoma detection results based on single-level graphs and \textbf{superpixel ensemble graphs (SEG)} with \textbf{learned weights} and  \textbf{geometric signals}. For each graph type, the best results are shown in bold. The best overall results are underlined.}\label{tab:Performance_Full_MML_Geometric}
		\centering
		\footnotesize
		\begin{tabular}{>{\centering\arraybackslash}m{2cm} >{\centering\arraybackslash}m{0.4cm}>{\centering\arraybackslash}m{0.85cm}>{\centering\arraybackslash}m{0.85cm}>{\centering\arraybackslash}m{0.85cm}>{\centering\arraybackslash}m{0.85cm}>{\centering\arraybackslash}m{0.85cm}>{\centering\arraybackslash}m{0.85cm}>{\centering\arraybackslash}m{0.85cm}>{\centering\arraybackslash}m{0.85cm}}
			\hline
			Graph level&	&	SVM   & KNN & QDA & RF & GB & AB  & MLP\\
			\hline
			Single graph&	AC	&	96.21	&96.79	&95.34	&	\textbf{97.96}&90.96&	96.50&	89.50	\\
			(n=20)	&	Spec&95.08&	93.13	&97.54	&\underline{\textbf{100.00}}&	99.18	&97.54&	98.36	\\
			&Sens	&97.27	&\textbf{99.01}	&94.12&	97.74	&97.29&	94.57	&95.48	\\
			& AUC& 98.05 & 98.48 & 94.88 & 98.35 & 90.02 & 97.20 &\textbf{ 98.90} \\ 
			
			\hline 
			
			Single graph&	AC	&	\textbf{98.38}	&97.08&	92.42&	98.71&	98.54&	95.34&	91.55	\\
			(n=40)	&	Spec&99.18	&93.13&	\underline{\textbf{100.00}}&	99.18&	99.18	&\underline{\textbf{100.00}}&	98.36	\\
			&Sens	&97.55&	99.52	&89.59	&\textbf{99.55}&	98.19	&94.57	&97.74	\\
			& AUC& 98.81 & 99.05 & 95.53 & \underline{\textbf{99.46}} & 91.63 & 98.48 & 99.33 \\ 
			
			\hline 
			
			Single graph&	AC	&	97.08&	\textbf{97.96}	&95.04&	95.34	&97.38	&95.92	&92.13	\\
			(n=60)	&	Spec&95.90&	94.57	&98.36	&95.08	&95.90&\textbf{99.08}	&98.36	\\
			&Sens	&97.55	&\underline{\textbf{100.00}}	&92.31&	99.55	&98.64	&97.29&	97.29	\\
			& AUC& 98.25 & \textbf{98.81} & 91.83 & 98.13 & 92.07 & 98.33 & 98.39 \\ 
			
			\hline 
			
			Single graph&	AC	&	\textbf{98.24}	&96.50	&93.00&	97.96&	94.96&	96.50	&92.13	\\
			(n=80)	&	Spec&98.36	&94.04	&99.18	&97.54	&\underline{\textbf{100.00}}	&93.96	&98.36	\\
			&Sens	&\textbf{99.55}	&99.21	&89.14	&98.35	&97.74&	95.02&	98.19	\\
			& AUC& 98.53 &\textbf{ 98.86} & 95.21 & 98.50 & 91.15 & 98.79 & 97.17 \\ 
			
			\hline 
			
			Single graph&	AC	&	\textbf{98.51}&	97.08&	87.17&	97.38&	97.24&	95.34	&93.29	\\
			(n=100)	&	Spec&96.72&	92.42	&98.18	&90.10	&99.10&	98.13&	\textbf{99.18}	\\
			&Sens	&97.26	&86.28&	94.12&	\textbf{99.55}&	99.00	&99.10	&98.64	\\
			& AUC& 98.77 & 98.40 & 93.13 & 98.13 & 88.82 & \textbf{99.16} & 97.90 \\
			\hline 
			
			Ensemble graph&	AC	&	98.54	&98.25&	94.67&	\underline{\textbf{98.57}}&	99.42&	95.34	&96.21	\\
			(fused n)	&	Spec&98.36&	95.31	&92.62	&\textbf{99.24}&	99.11	&99.18&	98.36	\\
			&Sens	&97.40	&\textbf{98.12}	&90.50&	97.25	&96.68	&95.93	&94.12 \\ 
			& AUC& 91.90 & 89.19 & 98.40 & 92.47 & 92.70 & \textbf{99.18} & 97.93 \\ 
			\hline 
			
		\end{tabular}
	\end{table*}
	
	
	\begin{table*}[h!]
		\centering
		\caption{Melanoma detection results based on single-level graphs and \textbf{superpixel ensemble graphs (SEG)} with \textbf{learned weights} and  \textbf{texture signals}. For each graph type, the best results are shown in bold. The best overall results are underlined.}\label{tab:Performance_Full_MML_Texture}
		\centering
		\footnotesize
		\begin{tabular}{>{\centering\arraybackslash}m{2cm} >{\centering\arraybackslash}m{0.4cm}>{\centering\arraybackslash}m{0.85cm}>{\centering\arraybackslash}m{0.85cm}>{\centering\arraybackslash}m{0.85cm}>{\centering\arraybackslash}m{0.85cm}>{\centering\arraybackslash}m{0.85cm}>{\centering\arraybackslash}m{0.85cm}>{\centering\arraybackslash}m{0.85cm}>{\centering\arraybackslash}m{0.85cm}}
			\hline
			Graph level&	&	SVM   & KNN & QDA & RF & GB & AB  & MLP\\
			\hline
			
			Single graph &	AC	&\textbf{98.23}	&95.92	&87.84	&90.35	&96.79&	96.79	&91.25\\	
			(n=20)&	Spec&93.77	&96.06	&96.07	&89.77	&\textbf{98.78}	&94.73	&98.36	\\
			&Sens	&95.93	&\textbf{99.52}&	82.81	&92.92	&97.29	&92.76&	90.05	\\
			& AUC& 95.97 & \textbf{99.32} & 98.17 & 91.72 & 98.94 & 96.37 & 99.28 \\ 
			\hline
			
			Single graph&	AC	&97.38&	96.79&	91.55&	90.64&	96.21&	\textbf{97.67}&	93.00	\\
			(n=40)&	Spec&\underline{\textbf{100.00}}&	93.13	&98.36&	99.18&	97.54&	98.36&	98.36	\\
			&Sens	&95.45&	\underline{\textbf{100.00}}&	87.78	&97.74&	96.38&	91.86&	91.86	\\
			& AUC& 99.14 & 97.46 & 96.85 & \textbf{99.12} & 90.64 & 97.36 & 95.64 \\ 
			\hline
			
			Single graph &	AC	&95.92	&\textbf{97.67}	&90.96&	90.94	&97.08	&96.79	&93.00	\\
			(n=60)&	Spec&91.80&	97.58	&97.54&	99.18	&\underline{\textbf{100.00}}&\underline{\textbf{100.00}}&	98.36	\\
			&Sens	&94.57	&\textbf{99.53}&	89.59	&91.54	&96.83	&92.76&	90.95	\\
			&AUC & 98.78 & 98.62 & 93.91 & \textbf{98.81} & 91.51 & 98.79 & 98.81 \\ 
			\hline
			
			Single graph &	AC	&88.60	&96.50&	94.17	&90.35	&\textbf{96.10}&	95.04&	90.67	\\
			(n=80)&	Spec&90.34&	91.04	&97.54	&97.62& 95.08 &97.50&\textbf{98.36}	\\
			&Sens	&89.82&\underline{\textbf{100.00}}&	95.02&	96.88&	96.38	&90.50	&88.69	\\
			&AUC & 93.14 & 98.94 & 93.65 & \textbf{98.02} & 90.68 & 98.97 & 96.37 \\ 
			\hline
			
			Single graph &	AC	&96.76	&96.50&	88.63&	90.64&	\textbf{97.38}&	95.92	&92.13	\\
			(n=100)&	Spec&95.08&	91.67&	94.71&	97.18	&95.08	&97.54	&\textbf{99.41}	\\
			&Sens	&97.29&\underline{	\textbf{100.00}}&	85.07&	83.26	&96.83	&91.40	&90.05	\\
			&AUC & \textbf{98.94} & 98.75 & 93.00 & 97.72 & 88.97 & 98.42 & 98.85 \\
			\hline
			
			Ensemble graph &	AC	&98.54&	97.96	&92.67&\underline{	\textbf{99.00}}&	98.76	&94.46&	93.88	\\
			(fused n)&	Spec&86.07&	96.00	&98.11&\underline{\textbf{100.00}}&	99.18&	99.18&	98.36	\\
			&Sens	&97.55	&\textbf{99.08}&	89.54&	98.64&	98.64	&97.74	&95.93	 \\
			& AUC& 92.41 & 97.99 & 97.37 & 93.16 & \underline{\textbf{99.59}} & 97.96 & 96.94 \\ 
			\hline

		\end{tabular}
	\end{table*}
	

	\begin{table*}[h!]
		\centering
		\caption{Melanoma detection results based on single-level graphs and \textbf{superpixel ensemble graphs (SEG)} with \textbf{learned weights} and \textbf{color signals}. For each graph type, the best results are shown in bold. The best overall results are underlined.}\label{tab:Performance_Full_MML_Color}
		\centering
		\footnotesize
		\begin{tabular}{>{\centering\arraybackslash}m{2cm} >{\centering\arraybackslash}m{0.4cm}>{\centering\arraybackslash}m{0.85cm}>{\centering\arraybackslash}m{0.85cm}>{\centering\arraybackslash}m{0.85cm}>{\centering\arraybackslash}m{0.85cm}>{\centering\arraybackslash}m{0.85cm}>{\centering\arraybackslash}m{0.85cm}>{\centering\arraybackslash}m{0.85cm}>{\centering\arraybackslash}m{0.85cm}}
			\hline
			Graph level&	&	SVM   & KNN & QDA & RF & GB & AB  & MLP\\
			\hline
			
			Single graph&	AC	&	\textbf{97.96}&	95.92&	94.75&	97.96&	97.67&	95.34&	92.13	\\
			(n=20)&	Spec&\textbf{	98.36}&	90.91&	96.72&	96.72&	98.36&	97.54&	\textbf{98.36}	\\
			&Sens	&	97.74&\textbf{	99.54}&	91.86&	99.10&	97.74&	94.57&	88.69\\	
			&AUC & \textbf{98.22} & 95.30 & 94.89 & 98.06 & 91.83 & 96.89 & 96.76 \\ 
			
			\hline
			
			Single graph	&	AC&\textbf{99.13}&	97.08&	91.84&	97.67&	97.67&	96.50&	93.29	\\
			(n=40)	&	Spec&98.00&	94.44&	98.36&	97.54&	98.36&	98.36 & \underline{\textbf{100.00}}	\\
			&	Sens&\underline{\textbf{100.00}}&99.53&	84.62&	99.55&	99.10&	93.67&	89.59	\\
			& AUC& \textbf{99.09} & 99.65 & 90.20 & 99.00 & 97.49 & 96.63 & 92.57 \\ 
			
			\hline 	
			
			Single graph	&	AC&95.34&	93.59&	94.17&	96.21&\textbf{98.25}&	93.88&	92.13	\\
			(n=60)	&	Spec&96.72&	86.23&	95.90&	94.26&	96.72&	97.54&	\textbf{99.18}	\\
			&	Sens&95.93&	99.01&	93.21&\textbf{99.09}&	96.38&	91.86&	95.69	\\ 
			& AUC& 98.20 & 98.84 & 93.83 & 98.91 & 89.19 & 98.86 & \textbf{98.99} \\ 
			
			\hline  
			Single graph&AC&	\textbf{97.96 }& 96.79 & 97.08 &95.34 &97.38 & 	96.50  & 	93.29	\\
			(n=80)&Spec&	95.90&	93.02&	98.36&	95.08&	98.36 & 	98.36 &\textbf{96.12}	\\
			&Sens&	\underline{\textbf{99.15}}	&99.07&	94.12&	96.47&	98.19&	92.76&	90.05	\\
			&AUC & 97.27 & \textbf{99.26} & 97.59 & 98.31 & 91.67 & 97.60 & 98.80 \\ 
			
			\hline 
			
			Single graph	&	AC&\textbf{97.67}&	95.34&	95.63&	96.21&	97.38&	93.59&	92.42\\
			(n=100)	&	Spec&98.36&	93.44&	95.90&	94.26&	95.90	&96.72&\textbf{	99.18}	\\
			&	Sens& 	97.74&	96.80&\textbf{	99.10}&	89.61&	97.29&	93.21&	88.69\\
			&AUC & \textbf{99.10} & 97.65 & 94.87 & 98.46 & 91.64 & 98.60 & 99.47 \\
			\hline 
			
			Ensemble  graph&AC& 98.70 & 98.05 &  96.42& \underline{\textbf{99.24}}&   98.70&   96.1039 &  94.80\\
			(fused n)&Spec&95.40  & 93.54 & \underline{\textbf{100.00}} & 98.97& 99.51 & 98.06 &91.30\\
			&Sens&97.69 & \underline{\textbf{100.00} } & 95.02 &  99.55 &  98.19 &  95.02   &90.47\\	
			& AUC& 97.94 & 98.24 & 98.87 & 92.38 & \underline{\textbf{99.31}} & 99.07 & 94.90 \\
			\hline 		
		\end{tabular}
	\end{table*}

	
	
	\begin{table*}[h!]
		\centering
		\caption{Summary of melanoma detection results using different multi-level superpixel ensemble graph (SEG) architectures with handcrafted and learned weights. For each weight assignment scheme, the best results are shown in bold. The best overall results are underlined.}\label{tab:Comparison_Results_diffent_LearnedGraphs}
		\centering
		\footnotesize
		\begin{tabular}{>{\centering\arraybackslash}m{1.5cm} >{\centering\arraybackslash}m{0.65cm}>{\centering\arraybackslash}m{1cm}>{\centering\arraybackslash}m{1cm}>{\centering\arraybackslash}m{1cm}}
			\hline
			& &Geometric &Texture	&Color\\
			\hline
			&AC&98.12 & 98.41	&\textbf{99.01} \\
			
			Handcrafted  &Spec&98.94 & \underline{\textbf{100.00}}	& \underline{\textbf{100.00}}\\
			
			weights  & Sens& 97.37 & 98.57	& \textbf{99.13}\\
			
			& AUC& 98.86 & 98.97 	& \textbf{99.10}\\
			\hline     
			&AC&98.71 & 99.00	& \underline{\textbf{99.24}}\\
			
			Learned  & Spec& 99.24 & \underline{\textbf{100.00}}& \underline{\textbf{100.00}} \\
			
			weights  & Sens & 98.12  & 99.08& \underline{\textbf{100.00}} \\
			& AUC& 99.18  & \underline{\textbf{99.59}}	& 99.31\\
			\hline 		
		\end{tabular}
	\end{table*}
	\section{Melanoma Detection System} \label{sec:Learning_Algorithm_Sec}
	The proposed melanoma detection system of this paper combines both conventional and graph signal features in the time and frequency domains. As indicated above, a multi-level superpixel graphs $G_k=\langle V_k,\varepsilon_k,W_k \rangle$ are constructd in handcradted and learned edge weighting approaches. We also extract conventional features from each dermoscopic image. In other words, the features employed in the proposed detection system can be categorized into three groups.
	
	\textbf{Time-domain graph features}. For each graph level $G_k$, we compute several vertex-domain features. This involves both local graph features (local efficiency (LE), local clustering coefficient (LCC), nodal strength (NS), nodal betweenness centrality (NBC), closeness centrality (CC) and eccentricity (Ecc)) and global graph features (characteristic path length (CPL), global 
	efficiency (GE), global clustering coefficient (GCC), 
	density (D) and global assortativity (GA)) \citep{stam2007graph}. These features are computed for nodal signals of the color, geometric, and texture types. The number of the time-domain graph features for each graph level is $F_1=n\times6+5$ features, where $n$ is the number of nodes in that level.
	
	\textbf{Frequency-domain graph features}. As mentioned above, the GFT of $G_k$ is computed in color, geometric and texture structures of $W_k$ \citep{huang2016graph}. The frequency-domain graph features implemented in the proposed system are energy, power, entropy and amplitude. The total number of the frequency-domain graph features is $F_2=n+4$ features for each graph level.
	
	\textbf{Conventional features}. Conventional color, geometric, and texture features \citep{oliveira2017skin} are extracted directly from the dermosopic images. The total count of these features is  $F_3=394$.
	
	The above-mentioned features are computed for each graph level, and are used to train single-graph-level classifiers. The features can be concatenated across multiple levels and used to train classifiers based on multi-level graph features. 
	
	To ensure uniform feature scaling, all extracted features are normalized to lie in the interval $[0,1]$. Also,  feature selection is performed using a logistic regression model with $L1$ regularization to identify the most important $150$ features. A 10-fold cross-validation scheme is applied to ensure stability and generalizability.
	
	\section{Experimental Results and Discussion}
	\label{sec:Results}
	\subsection{Experimental Setup}
	In this work, we evaluated the performance of multi-level superpixel graphs constructed using the augmented ISIC$2017$ dataset (Section \ref{sec:DataSet}), which includes a diverse set of skin lesion images. A stratified 10-fold cross-validation scheme was employed for all experiments.
	
	The images were analyzed on three distinct categories of image features: geometry, texture, and color. For each feature category, superpixel graphs were generated with $20, 40, 60, 80, $ and $100$ nodes, respectively. These graphs were designed to capture the complex patterns within the skin lesion images at different levels of granularity based on each of the three types of image features. 
	
	The weights of the graph edge were either computed using a Gaussian kernel (Eq. \ref{eq:Gauss_function}) or learned using the majorization-minimization learning (MML) algorithm. All constructed graphs are fully connected graphs with no self-loops. 
	
	
	Multi-level graph representations were created by concatenating features from five single-level superpixel graph representations with $20, 40, 60, 80,$ and $ 100$ nodes, respectively. These representations were based on geometric, texture, or color features.  
	
	For each of the single-level and multi-level graph representations, the top $150$ features were selected using the logistic regression model with $L1$ regularization. The training features were thus used to train the following classifiers: support vector machines (SVM), k-nearest neighbors (KNN), multilayer perceptrons (MLP), random forests (RF), gradient boosting (GB), and AdaBoost (AB).
	
	The performance of each classifier was evaluated using metrics of accuracy (AC), specificity (Spec), sensitivity (Sens), and area under the ROC curve (AUC). 
	
	Detailed experimental results are provided in the next few sections. In particular, Tables \ref{tab:Performance_Full_gauss_Geometric}, \ref{tab:Performance_Full_gauss_Texture}, and \ref{tab:Performance_Full_gauss_Color} present the classification results for single-level graphs and superpixel ensemble graphs (SEG) with handcrafted edge weights. Corresponding results for graphs with MML-based structure learning are given in Tables \ref{tab:Performance_Full_MML_Geometric}, \ref{tab:Performance_Full_MML_Texture}, and \ref{tab:Performance_Full_MML_Color}. As well, Table \ref{tab:Comparison_Results_diffent_LearnedGraphs} summarizes the results based on both handcrafted and learned edge weights across geometric, texture, and color graph signals. Tables \ref{tab:Performance_Full_MML_Geometric_SHG}, \ref{tab:Performance_Full_MML_Texture_SHG}, and \ref{tab:Performance_Full_MML_Color_SHG} show corresponding graph-learning-based results for superpixel hierarchy graphs (SHG). Table \ref{tab:SOTA_Comparison} further compares the results of our proposed approach versus other state-of-the-art approaches, highlighting the similarities and differences in performance.
	
	
	\subsection{Results of Melanoma Detection using Superpixel Ensemble Graphs with Handcrafted Edge Weights}
	\label{Detection_results_for_the_Gaussian_weighted_multilevel_superpixel_graphs}
	In this section, we present the results of our melanoma detection approach using superpixel ensemble graphs (SEG) with handcrafted edge weights and nodal signals of gemoetric, texture, or color types. 
	\subsubsection{Geometric Nodal Signals}
	Table \ref{tab:Performance_Full_gauss_Geometric} shows the melanoma detection results for these graphs with geometric nodal signals and a 10-fold cross-validation scheme on the augmented ISIC$2017$ dataset at different levels. 
	The overall detection accuracy clearly varies with the number of graph nodes. For the single-level graph representations, the SVM and GB classifiers achieved accuracies of $98.04\%$ and $98.00\%$ at graph levels of 40 and 80, respectively. Moreover, the multi-level graph representation led to a better accuracy of $98.10\%$.\\ 
	The results show that the sensitivity frequently improved with higher node counts for single-level graphs and with the adoption of multi-level SEGs. This is particularly evident for the KNN classifier which mostly achieved the best sensitivity values among all classifiers, and reached a sensitivity of $98.53\%$ for a single-level graph representation with $40$ nodes.\\
	The results show that the GB classifier generally achieved the highest specificity of $100.00\%$ for the single-level graph representation with $60$ nodes. 
	
	The performance of multi-level graphs also warrants attention. The integration of features from all node configurations provided a comprehensive representation, resulting in superior performance metrics. For instance, the multi-level graph configuration yielded the highest accuracy of $98.10\%$, with specificity and sensitivity scores of $98.94\%$ and $97.37\%$ respectively, indicating its effectiveness in leveraging diverse feature sets.
	
	\subsubsection{Texture Nodal Signals}
	Table \ref{tab:Performance_Full_gauss_Texture} shows the melanoma detection results using superpixel ensemble graphs (SEG) with handcrafted edge weights and texture nodal signals. The classifiers exhibited variable performance based on the number of nodes in the constructed graphs. Overall, the classifiers achieved peak performance at different node counts, showing notable improvements with the multi-level graph approach.
	
	In particular, the results show that the SVM classifier mostly maintained high accuracy across most single-level graphs, achieving a remarkable $97.83\%$ for $100$ nodes. The multi-level graph enhances the accuracy further to $98.41\%$. Also, sensitivity demonstrated improvements with higher node counts. The KNN classifier  excelled at almost all single-level graphs, reaching a value of $99.03\%$ for the $20$-node graph. Moreover, a perfect specificity ($10
	0.00\%$) was attained by the AB, RF, and SVM classifiers for multiple single-level graphs as well as the multi-level graph. This highlights the robustness of our approach in accurately identifying non-melanoma cases, thereby reducing the risk of false positives. 
	\subsubsection{Color Nodal Signals}
	Table \ref{tab:Performance_Full_gauss_Color} shows the melanoma detection results for superpixel ensemble graphs (SEG)  with handcrafted edge weights and color nodal signals. 
	High accuracies were achieved by multiple classifiers trained on different variants of single-level graphs. Still, the RF classifier trained on features of multi-level graphs attained the best overall performance metrics with an accuracy of $99.01\%$, a sensitivity of $99.13\%$, and a specificity of $100.00\%$. These are indeed the best results among all those reported for all graph architectures with handcrafted weights. Such results indicate the effectiveness of the multi-level graph approach in leveraging diverse feature sets, leading to remarkable melanoma detection performance.
	
	\subsection{Results of Melanoma Detection using Superpixel Ensemble Graphs with Learned Edge Weights}
	\label{Detection_results_for_the_automated_weighted_multilevel_superpixel_graphs}
	In this section, we examine the performance of our melanoma detection approach where handcrafted edge weights are replaced by data-driven weights learned using the majorization-minimization learning (MML) framework (Eq. \ref{eq:min_Fn}). We reevaluated the performance of the seven classifiers trained on features of single-level and multi-level graph models for geometric, texture, and color nodal signals.  
	
	\subsubsection{Geometric Nodal Signals}
	
	Table \ref{tab:Performance_Full_MML_Geometric} shows the melanoma detection results for classifiers trained on features of geometric graph signals with learned edge weights. The SVM and RF classifiers achieved the best accuracies with different graph levels. Specifically, the RF classifier trained on multi-level graph features scored the highest accuracy of $98.57\%$. The KNN and RF classifiers gave the best sensitivities, where a perfect sensitivity of $100\%$ was attained by the KNN classifier trained on 60-node graph features. As well, a perfect specificity of $100\%$ was reached using single-level graphs of 20, 40, and 80 nodes.
	It's noteworthy that the graph-theoretic models with learned weights achieved better performance metrics than models with handcrafted edge weights (See Table \ref{tab:Comparison_Results_diffent_LearnedGraphs}).
	
	\subsubsection{Texture Nodal Signals}
	
	For classifiers trained on  features of texture graph signals with learned edge weights, the melanoma detection results are presented in Table \ref{tab:Performance_Full_MML_Texture}. The best accuracy among single-level graph models was attained by the 20-node graph, while the multi-level graph led to the best overall accuracy of $99.00\%$. The best sensitivity results were consistently attained by the KNN classifier across all graph models, reaching $100\%$ for most single-level graphs. A perfect specificity of $100\%$ was obtained by multiple classifiers trained on single-level and multi-level graph features.
	In comparison to graph-theoretic models with handcrafted edge weights, the ones with learned edge weights demonstrated enhanced performance metrics for the texture nodal signals (See Table \ref{tab:Comparison_Results_diffent_LearnedGraphs}).
	
	\subsubsection{Color Nodal Signals}
	Table \ref{tab:Performance_Full_MML_Color} shows the melanoma detection results for classifiers trained on features of color graph signals with learned edge weights. The SVM classifier gave the best accuracy for four single-level graph models, while the RF classifier achieved the best overall accuracy of $99.24\%$ with multi-level graphs. This accuracy is indeed the best result attained overall all classifiers, graph models, and nodal signal types (See also Table \ref{tab:Comparison_Results_diffent_LearnedGraphs}).
	This particular result shows the significance of the proposed multi-level graph representation along with graph learning and color nodal signals. 
	\begin{table*}[h!]
		\centering
		\caption{Melanoma detection results based on single-level graphs and \textbf{superpixel hierarchy graphs (SHG)} with \textbf{learned weights} and  \textbf{geometric signals}. For each graph type, the best results are shown in bold. The best overall results are underlined.}\label{tab:Performance_Full_MML_Geometric_SHG}
		\centering
		\footnotesize
		\begin{tabular}{>{\centering\arraybackslash}m{2cm} >{\centering\arraybackslash}m{0.4cm}>{\centering\arraybackslash}m{0.85cm}>{\centering\arraybackslash}m{0.85cm}>{\centering\arraybackslash}m{0.85cm}>{\centering\arraybackslash}m{0.85cm}>{\centering\arraybackslash}m{0.85cm}>{\centering\arraybackslash}m{0.85cm}>{\centering\arraybackslash}m{0.85cm}>{\centering\arraybackslash}m{0.85cm}}
			\hline
			Graph level&	&	SVM   & KNN & QDA & RF & GB & AB  & MLP\\
			\hline
			Single graph&	AC	& \textbf{97.95} & 94.31 & 93.46 & 97.65 & 96.71 & 93.56 & 92.68\\
			(n=5)&	Spec& \textbf{98.95} & 88.27 & 98.25 & 97.87 & 97.11 & 98.65 & 98.13\\ 
			&Sens	&97.17 & 94.80.0 & 97.21 & 97.64 & 94.24 & 88.82 & \textbf{98.74}\\
			& AUC& 96.51 & \textbf{97.97} & 97.06 & 94.87& 95.94 & 97.75 & 97.01 \\ 
			
			\hline 
			
			Single graph&	AC	& \textbf{98.23} & 96.41 & 73.05 & 98.2 & 97.60 & 95.51 & 94.01\\
			(n=10)&	Spec&97.9 & 92.31 & 98.62 & 90.91 & 90.91 & \textbf{98.00} & 97.2\\
			&Sens	&97.91 & \underline{\textbf{99.04}} & 92.16 & 98.95 & 96.86 & 95.29 & 91.10\\
			& AUC& 91.96 & 93.72 & \textbf{95.66} & 93.44 & 93.20 & 92.57 & 91.62  \\ 
			
			\hline 
			
			Single graph&	AC	&	 97.06 & 97.90 &91.65 & \textbf{98.02} & 97.01 & 96.71 & 93.11\\
			(n=20)&	Spec&\underline{\textbf{99.03}} & 95.33 & 97.9 & 89.51 & 90.91 & 90.91 & 98.6\\
			&Sens	&97.38 & 91.61 & 76.44 & 98.43 & \textbf{98.95} & 94.76 & 89.01\\
			& AUC& 92.09 & 93.95 & 96.66 & 93.96 & 94.48 & 93.48 & \textbf{97.58} \\ 
			
			\hline 
			
			Single graph&	AC	&	 98.2 & 95.51 & 89.75 & \textbf{98.8} & 96.41 & 97.31 & 93.11\\
			(n=40)&	Spec& \textbf{98.60} & 94.7 & 92.31 & 89.51 & 91.51 & 90.21 & 97.92\\ 
			&Sens	&96.86 & 92.81 & 95.29 & \textbf{98.95} & 96.86 & 97.38 & 97.38\\
			& AUC& 91.93 & 93.34 & \textbf{97.52} & 93.44 & 93.75 & 92.77 & 89.96 \\ 
			
			\hline 
			
			Single graph&	AC	& 98.42 & 98.08 & 82.63 & \textbf{98.50} & 98.11 & 97.90 & 94.61\\
			(n=80)&	Spec&98.15 & 97.28 & \textbf{98.61} & 89.51 & 91.61 & 90.21 & 98.36\\
			&Sens	& \textbf{98.95} & 92.41 & 89.44 & 98.73 & 97.91 & 93.16 & 97.91\\
			& AUC& 93.23 & 94.11 & 97.61 & 94.29 & 94.24 & 92.87 & \textbf{98.17} \\
			\hline 
			
			Hierarchy &	AC	& 95.85 & 96.11 & 96.71 & \underline{\textbf{98.94}} & 95.81 & 92.51 & 89.97\\
			graph&	Spec&97.08 & 91.67 & \textbf{99.00} & 97.54 & 94.72 & 93.16 & 97.61\\
			(fused n)&Sens	&94.74 & \textbf{98.33} & 95.03 & 97.95 & 92.65 & 89.22 & 82.98\\
			& AUC& 97.48 & 95.74 & \underline{\textbf{98.57}} & 98.01 & 91.07 & 96.13 & 96.43 \\ 
			\hline 
			
		\end{tabular}
	\end{table*}
	
	\begin{table*}[h!]
		\centering
		\caption{Melanoma detection results based on single-level graphs and \textbf{superpixel hierarchy graphs (SHG)} with \textbf{learned weights} and  \textbf{texture signals}. For each graph type, the best results are shown in bold. The best overall results are underlined.}\label{tab:Performance_Full_MML_Texture_SHG}
		\centering
		\footnotesize
		\begin{tabular}{>{\centering\arraybackslash}m{2cm} >{\centering\arraybackslash}m{0.4cm}>{\centering\arraybackslash}m{0.85cm}>{\centering\arraybackslash}m{0.85cm}>{\centering\arraybackslash}m{0.85cm}>{\centering\arraybackslash}m{0.85cm}>{\centering\arraybackslash}m{0.85cm}>{\centering\arraybackslash}m{0.85cm}>{\centering\arraybackslash}m{0.85cm}>{\centering\arraybackslash}m{0.85cm}}
			\hline
			Graph level&	&	SVM   & KNN & QDA & RF & GB & AB  & MLP\\
			\hline
			
			Single graph&	AC	&  96.33 & 95.75 & 93.78 & 93.24 & \textbf{96.53} & 93.63 & 94.45\\
			(n=5)	&	Spec&91.44 & 91.02 & \underline{\textbf{98.65}} & 93.14 & 92.89 & 96.69 & 98.21\\
			&Sens	& 93.56 & 91.72 & 96.96 & 94.42 & 93.9 & 89.81 & \textbf{97.99}\\
			& AUC& 92.73 & \textbf{97.61} & 95.33 & 96.99 & 97.38 & 98.74 & 97.11 \\ 
			\hline
			
			Single graph&	AC	& 94.98 & \textbf{96.72} & 94.85 & 93.63 & 90.54 & 93.24 & 89.38\\
			(n=10)	&	Spec& 97.02 & 92.92 & \textbf{98.33} & 97.10 & 96.89 & 93.09 & 98.21\\
			&Sens	&\textbf{97.97 }& 95.58 & 96.96 & 97.63 & 96.61 & 97.14 & 92.71\\ 
			& AUC& 97.81 & 96.28 & 95.95 & \textbf{97.98} & 94.55 & 95.24 & 96.42 \\
			
			\hline 
			
			Single graph&	AC	&\textbf{97.07} & 96.53 & 92.28 & 94.92 & 96.72 & 94.79 & 90.35\\
			(n=20)	&	Spec&\textbf{98.25} & 92.53 & 98.12 & 95.07 & 94.32 & 97.72 & 93.71\\
			&Sens	& 97.97 & 95.41 & 97.86 & 97.14 & 94.24 & 90.85 & \textbf{98.31}\\
			& AUC& 95.6 & \textbf{97.71} & 94.98 & 93.45 & 90.21 & 94.71 & 95.42 \\ 
			\hline  
			
			Single graph&	AC	&\underline{\textbf{98.46}} & 95.17 & 94.98 & 95.75 & 97.88 & 94.4 & 91.51\\ 
			(n=40)	&	Spec& \textbf{97.10} &92.89 & 98.65 & 96.21 & 95.37 & 94.79 & 98.65\\ 
			&Sens	&97.97 & 94.22 & 96.96 & 92.54 & 96.27 & 90.17 & \textbf{98.05} \\
			& AUC& 94.69 & 92.55 & 95.48 & 92.74 & 95.51 & 94.12 & \textbf{97.23} \\ 
			\hline 
			
			Single graph&	AC	&  \textbf{98.07} & 92.66 & 97.88 & 96.33 & 97.3 & 93.44 & 89.10\\
			(n=80)	&	Spec& \textbf{98.21} & 97.44 & 98.01 & 96.25 & 96.78 &  93.54 & 97.81 \\ 
			&Sens	&97.97 & 98.74 & 96.62 & 93.56 & 95.25 & \textbf{98.47} & 97.64\\ 
			& AUC& 89.06 & 97.10 & 96.95 & 97.09 & 98.03 & \textbf{97.59} & 97.33 \\
			\hline 
			
			Hierarchy &	AC	&  \textbf{98.07} & 97.40 & 96.72 & 97.97 & 97.3 & 94.79 & 90.15\\
			graph	&	Spec&97.31 & \textbf{98.49} & 98.21 & 97.14 & 98.10 & 98.32 & 98.10\\
			(fused n)&Sens	&\underline{\textbf{98.94}} & 97.70 &  98.32 & 98.31 & 95.25 & 98.39 & 97.64\\ 
			& AUC&94.86 & 96.4 & 97.97 & \underline{\textbf{98.99}} & 94.86 & 93.79 & 91.15 \\ 
			\hline 
			
		\end{tabular}
	\end{table*}
	
	\begin{table*}[h!]
		\centering
		\caption{Melanoma detection results based on single-level graphs and \textbf{superpixel hierarchy graphs (SHG)} with \textbf{learned weights} and \textbf{color signals}. For each graph type, the best results are shown in bold. The best overall results are underlined.}\label{tab:Performance_Full_MML_Color_SHG}
		\centering
		\footnotesize
		\begin{tabular}{>{\centering\arraybackslash}m{2cm} >{\centering\arraybackslash}m{0.4cm}>{\centering\arraybackslash}m{0.85cm}>{\centering\arraybackslash}m{0.85cm}>{\centering\arraybackslash}m{0.85cm}>{\centering\arraybackslash}m{0.85cm}>{\centering\arraybackslash}m{0.85cm}>{\centering\arraybackslash}m{0.85cm}>{\centering\arraybackslash}m{0.85cm}>{\centering\arraybackslash}m{0.85cm}}
			\hline
			Graph level&	&	SVM   & KNN & QDA & RF & GB & AB  & MLP\\
			\hline
			Single graph &	AC	&\textbf{97.86} & 93.13 & 95.46 & 96.02 & 96.77 & 93.41 & 95.41\\
			(n=5)&	Spec&98.73 & 95.76 & 97.97 & 97.00 & 98.57& 98.66 & \underline{\textbf{98.98}}\\ 
			&Sens	&  \textbf{98.33} & 97.98 & 90.65 & 97.85 & 94.50 & 95.65 & 98.80\\
			& AUC& 94.99 & 96.63 & 97.49 & \textbf{97.01} & 95.00 & 96.93 & 96.41 \\
			\hline
			Single graph &	AC	& 96.86 & 92.29 & 92.93 & \textbf{98.02} & 96.49 & 93.69 & 89.69\\ 
			(n=10)&	Spec&99.66 & 94.29 & 98.31 & 93.90 & 97.32 & 97.66 & \textbf{98.85}\\
			&Sens	&\textbf{98.56} & 97.19 & 89.93 & 96.89 & 92.58 & 88.52 & 98.52\\
			& AUC& 96.86 & 97.26 & \textbf{98.02} & 93.97 & 97.78 & 94.97 & 91.92 \\
			\hline
			Single graph &	AC	& 97.86 & 93.27 & 88.98 & \textbf{98.07} & 96.91 & 94.44 & 89.83\\
			(n=20)&	Spec&96.95 & 93.22 & 98.31 & 96.54 & 98.27 & 97.33 & \textbf{98.64}\\
			&Sens	&\textbf{98.76} & 96.73 & 97.05 & 98.09 & 94.74 & 96.57 & 97.52\\
			& AUC& 97.35 & 96.52 & 97.56 & 97.22 & 93.94 & \textbf{97.90} & 97.01 \\
			\hline
			Single graph &	AC	& \textbf{97.58} & 93.41 & 89.76 & 99.30 & 96.35 & 92.57 & 94.99\\
			(n=40)&	Spec& \textbf{98.02} & 96.26 & 98.31 & 95.61 & 97.27 & 96.66 & 92.32\\
			&Sens	&97.28 & 96.67 & 96.09 & \textbf{98.63} & 93.78 & 94.32 & 98.28\\ 
			& AUC& 95.96 & \textbf{97.88} & 97.85 & 92.37 & 96.84 & 94.97 & 97.03 \\
			\hline
			Single graph &	AC	& \underline{\textbf{98.72}} & 96.77 & 92.23 & 98.31 & 96.63 & 92.85 & 90.41\\
			(n=80)&	Spec&98.12 & 92.77 & 98.31 & \textbf{98.34}& 96.70& 97.32 & 95.32\\ 
			&Sens	&\textbf{98.81} & 97.17 & 98.40 & 97.61 & 94.26 & 97.8 & 98.08\\
			& AUC& \underline{\textbf{98.33}} & 96.15 & 96.85 & 97.99 & 94.89 & 96.85 & 96.73 \\
			\hline
			Hierarchy  &	AC	& \textbf{98.10} & 92.99 & 93.07 & 97.96 & 97.48 & 93.41 & 87.38\\
			graph&	Spec& 98.11 & 90.71 & 98.64 & 98.24& 98.66 & \underline{\textbf{98.98}} & 98.31\\
			(fused n)&Sens	&98.32 & 98.84 & \underline{\textbf{99.08}} & 97.61 & 92.04 & 93.07 & 98.08\\
			& AUC& 97.13 & 97.71 & \textbf{97.93} & 97.01 & 98.09 & 96.82 & 94.51 \\
			\hline 		
		\end{tabular}
	\end{table*}
	
	\section{Discussion}
	\label{Sec:Discussion}
	
	The above-mentioned results show the melanoma detection performance improvements gained through employing multiple types of dermoscopic image characteristics,  multi-level graph features, and graph structure learning, and color information in dermoscopic image.  
	Furthermore, we discuss the performance differences between superpixel ensemble graphs and superpixel hierarchy graphs (\ref{subsec:Superpixel_ensembles versus_superpixel_hierarchies_for_melanoma_detection}), performance comparison based on the area under the ROC curve (\ref{subsec:ROC analysis}), feature significance analysis (\ref{subsec:Feature Level Contribution Analysis}), and effects of graph pruning on performance (\ref{subsec:Effect of graph thresholding (Edge Pruning) on melanoma detection Performance}).
	\subsection{Superpixel Ensembles versus Superpixel Hierarchies for Melanoma Detection}
	\label{subsec:Superpixel_ensembles versus_superpixel_hierarchies_for_melanoma_detection}
	We further investigated differences in melanoma detection performance between superpixel ensemble graphs (SEG) and superpixel hierarchical graphs (SHG). As mentioned in Section \ref{sec:Multi_Level_hierarchical_superpixel_graph}, there is no hierarchy among the superpixel map levels in a superpixel ensemble, while a superpixel hierarchy is constrained  by explicit parent-child relationships between superpixels at any two consecutive levels (See Fig. \ref{fig:Two_SPLevels}). We conducted experiments on melanoma detection using superpixel hierarchy graphs with learned weights. The detection results are shown for geometric features (Table \ref{tab:Performance_Full_MML_Geometric_SHG}), texture features (Table \ref{tab:Performance_Full_MML_Texture_SHG}), and color features (Table \ref{tab:Performance_Full_MML_Color_SHG}).
	Using geometric features, the best   accuracy for SHG features was $98.94\%$ for the RF classifier (Table \ref{tab:Performance_Full_MML_Geometric_SHG}), and this is slightly better than the best accuracy ($98.57\%$) obtained with the SEG features (Table \ref{tab:Performance_Full_MML_Geometric}). The best  sensitivity for the SEG-based detection ($100\%$ with a KNN) is slightly better than that of the SHG-based detection ($99.04\%$ with a 10-node single graph and a KNN classifier).

	Using texture features, the best accuracy for SHG features was $98.46\%$ for the SVM classifier (Table \ref{tab:Performance_Full_MML_Texture_SHG}), and this is slightly close to accuracy ($99.00\%$) obtained with the SEG features (Table \ref{tab:Performance_Full_MML_Texture}). The best  sensitivity for the SHG-based detection ($98.94\%$ with a SVM and hierarchy graph features) is slightly less than that obtained through SEG-based detection ($100.00\%$ with a 100-node single graph and a KNN classifier).
	 \begin{figure*}[!t]	
		 	\centering
		 	\includegraphics[width=\textwidth]{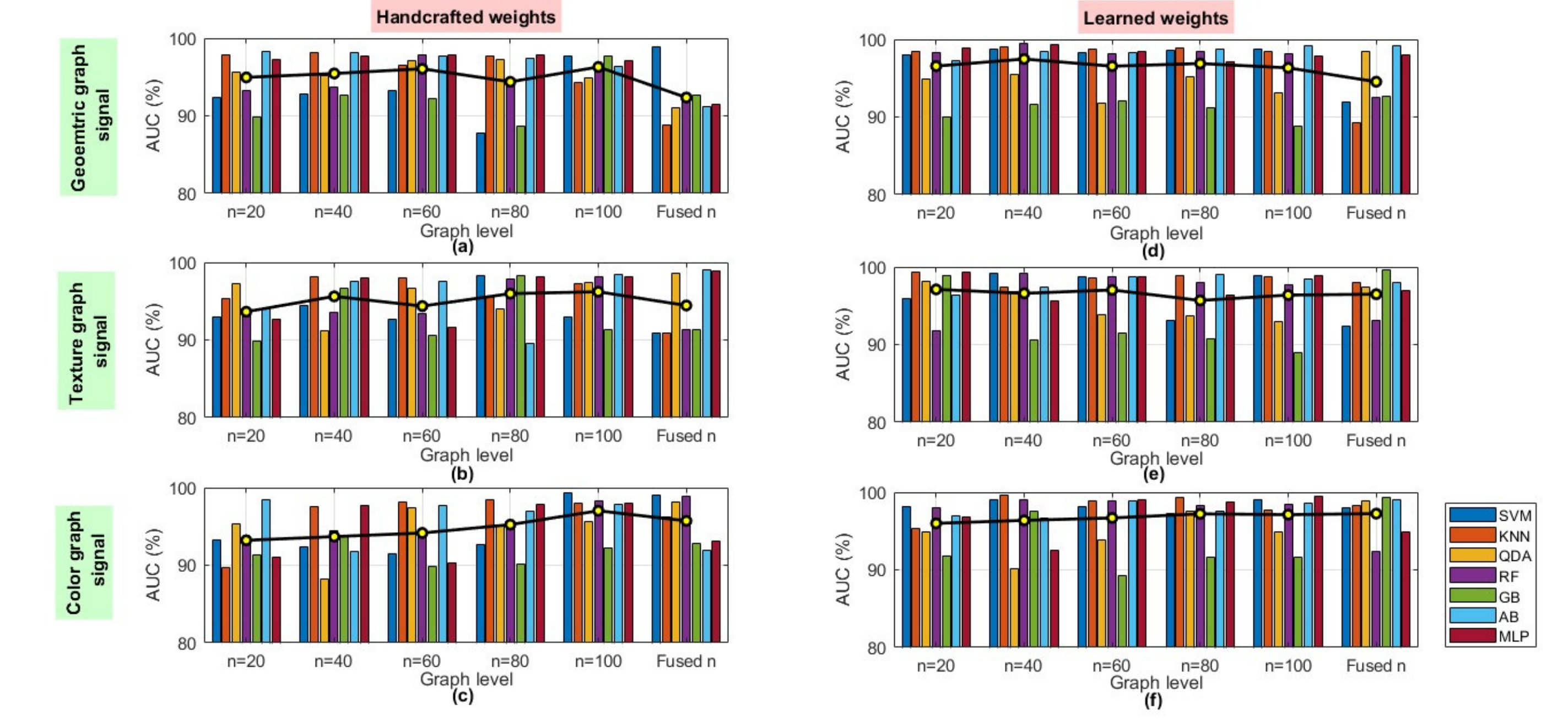} 
		 	\caption{Areas under the ROC curve (AUCs) for seven different classifiers trained on features of single-level graphs and superpixel ensemble graphs (SEG): (a) geometric graph
			 		signals and handcrafted weights, (b) texture graph signals and handcrafted weights, (c) color graph signals and handcrafted weights, (d) geometric graph
			 		signals and learned weights, (e) texture graph signals and learned weights, (f) color graph signals and learned weights.}
		 	\label{fig:AUC_BarChart_AllGraphs}
		 \end{figure*}
	
	
	\subsection{Overall Performance Comparison based on AUC}
	\label{subsec:ROC analysis}
	
	Based on superpixel ensemble graphs (SEG), Figure \ref{fig:AUC_BarChart_AllGraphs} shows bar charts for the AUC values computed for the seven classifiers, the three types of nodal signals, and the two methods of edge weight assignment. Figure \ref{fig:AUC_BarChart_AllGraphs}(a) shows the AUC values with handcrafted weights and geometric nodal signals. While classifiers based on features of single-level graphs achieve AUC values above 0.90, the multi-level graph leads to the best AUC with the SVM classifier (See also Table \ref{tab:Performance_Full_gauss_Geometric}). With texture nodal signals (Figure \ref{fig:AUC_BarChart_AllGraphs}(b)), the multi-level graphs still show superior performance (with the RF, AB, and MLP classifiers) compared to single-level graphs (See also Table \ref{tab:Performance_Full_gauss_Texture}). For color nodal signals (Figure \ref{fig:AUC_BarChart_AllGraphs}(c)), the SVM classifier had the best AUC with a 100-node graph, while the QDA classifier with multi-level graphs followed with a slightly smaller AUC value (See also Table \ref{tab:Performance_Full_gauss_Color}). 
	Corresponding AUC results based on learned edge weights are shown in Figures \ref{fig:AUC_BarChart_AllGraphs}(d-f) for the geometric, texture, and color nodal signals, respectively. The best AUC values for the geometric and color nodal signals are achieved by the RF and KNN, respectively, using 40-node single-level graphs. For the texture nodal signals, the GB classifier trained on the multi-level graph features led to the best AUC value ($99.59\%$) over all explored methods (See Table \ref{tab:Comparison_Results_diffent_LearnedGraphs}).  
	
	Overlaid on each of these bar plots is a mean AUC line that connects the average performance of all classifiers across different graph levels. This line serves as an important visual indicator of how well the feature representations generalize across classifiers. A rising or consistently high mean AUC line indicates feature robustness and classification performance stability, regardless of the specific classifier used. In our results, higher graph levels with more nodes generally exhibit higher average AUC values, reflecting the richer structural and signal information captured at those levels.
	This visualization facilitates both classifier-wise and level-wise performance interpretation and supports our fusion approach by illustrating the complementary strengths of different graph scales.
	
	
	 \begin{figure*}[!t]
		 	\centering
		 	\includegraphics[width=\textwidth]{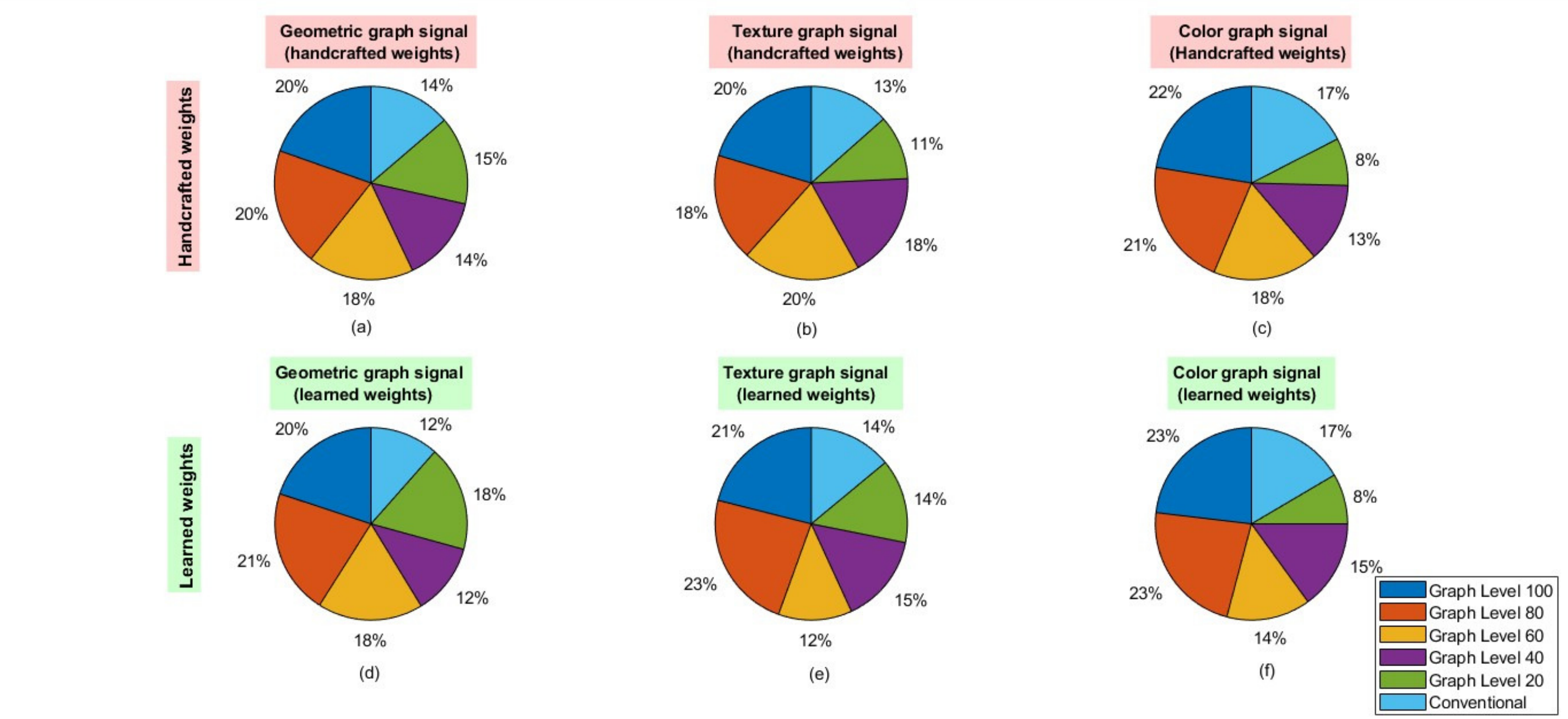} 
		 	\caption{{Distribution of selected features across five graph levels and conventional features used for superpixel ensemble graphs (SEG) constructed with handcrafted and learned weights: (a) geometric graph signals with handcrafted weights, (b) texture graph signals with handcrafted weights, (c) color graph signals with handcrafted weights, (d) geometric graph signals with learned weights, (e) texture graph signals with learned weights, (f) color graph signals with learned weights.}}
		 	\label{fig:Features_Contribution_PieChart}
		 \end{figure*}
	
	\subsection{Feature Contribution Analysis for Multi-Level Graph Representations}
	\label{subsec:Feature Level Contribution Analysis}
	Understanding the significance of each feature type helps to evaluate the role of multi-level graph representations in the detection of melanoma. We demonstrate the significance of the features extracted at different levels of superpixel ensemble graphs (SEG) via investigating the composition of the top $150$ features selected using logistic regression with $L1$ regularization \citep{ng2004feature}. Each possible configuration can be categorized along three dimensions. The first dimension is the type of the graph signal (geometric, texture, or color). The second dimension pertains to the method of assigning graph edge weights (handcrafted or learned edge weights). The third dimension is the feature type where features can be conventional features \citep{annaby2021melanoma} or one of five possible graph features (namely, those of single-level graphs with 20, 40, 60, 80, or 100 nodes, respectively). The extracted graph-theoretic features include graph Fourier transform (GFT) coefficients and vertex-domain features  derived at different graph levels $(20, 40, 60, 80,$ and $100$ nodes). The contribution of these features is analyzed by examining their selection frequency during the feature selection process for multi-level graph representations.
	
	For each combination of graph signal type and weight assignment method, Figure \ref{fig:Features_Contribution_PieChart} shows the contribution of the conventional features and the graph-theoretic features (from five graph levels) in the construction of the features of the proposed multi-level graphs. In particular, Figures \ref{fig:Features_Contribution_PieChart}(a-c) show feature distributions using a handcrafted methodology for edge weight assignment, along with signals of the geometric, texture, and color types, respectively. Similarly, Figures \ref{fig:Features_Contribution_PieChart}(d-f) show corresponding feature distributions using learned edge weights. 
	
	Clearly, all six feature types contribute significantly to the selected features (with the smallest contribution of $8\%$ for conventional features). Second, the dependence of graph signals on higher graph levels is more evident in the case of color signals than in the case of texture and geometric signals. Third, the feature distribution varies across the weight assignment dimension, where the schemes with learned weights tend to depend more on higher graph levels (i.e., graphs with more nodes). 
	
	\begin{table*}[h!]
		\centering
		\caption{Melanoma detection results for thresholded superpixel ensemble graphs (SEG) with learned weights and color signals. Graphs are pruned with three thresolds ($\tau\%=25\%,50\%,75\%$). For each threshold, the best results are shown in bold. The best overall results are underlined.}\label{tab:Performance_Threshold_MML_Color}
		\centering
		\footnotesize
		\begin{tabular}{>{\centering\arraybackslash}m{2cm} >{\centering\arraybackslash}m{0.4cm}>{\centering\arraybackslash}m{0.85cm}>{\centering\arraybackslash}m{0.85cm}>{\centering\arraybackslash}m{0.85cm}>{\centering\arraybackslash}m{0.85cm}>{\centering\arraybackslash}m{0.85cm}>{\centering\arraybackslash}m{0.85cm}>{\centering\arraybackslash}m{0.85cm}>{\centering\arraybackslash}m{0.85cm}}
			\hline
			Pruning threshold  ($\tau\%$)&	&	SVM   & KNN & QDA & RF & GB & AB  & MLP\\
			\hline
			
			&AC& 98.70 & 98.05 &  96.42& \underline{\textbf{99.24}}&   98.70&   96.1039 &  94.80\\
			$\tau$=0\% &Spec&95.40  & 93.54 & \underline{\textbf{100.00}} & 98.97& 99.51 & 98.06 &91.30\\
			&Sens&97.69 & \underline{\textbf{100.00} } & 95.02 &  99.55 &  98.19 &  95.02   &90.47\\	
			& AUC& 97.94 & 98.24 & 98.87 & 92.38 & \textbf{99.31} & 99.07 & 94.90 \\
			\hline	
			&	AC	&92.57	&93.86&	89.66	&\textbf{97.85}&	95.11	&90.44	&87.71		 	\\
			$\tau$=25\% &	Spec&	85.83&	87.30	&93.87&	\textbf{97.21}&	87.78&	91.61&	91.67\\
			&Sens	&93.12&	97.65&	90.87&	\textbf{98.14}&	96.83&	94.57&	93.67		 \\	
			& AUC& 92.12 & 96.06 & 92.00 &\textbf{ 98.99} & 93.07 & 98.61 & 97.99 \\ 
			\hline
			
			&	AC& 88.03	&93.86&	84.43&	\textbf{98.10}	&96.25	&92.15&	88.40	 	 	\\
			$\tau$=50\%&	Spec&	92.50&	91.38&	\textbf{98.97}&	97.22&	97.98&	98.61&	95.83		 		\\
			&	Sens &	97.11&	93.01&	91.40&	97.29&	\textbf{97.74}&	96.38&	95.02		\\
			& AUC& 89.91 & 97.36 & 96.97& \underline{\textbf{99.36}} & 98.89 & 89.96 & 98.03 \\ 	
			\hline 	
			
			&	AC&89.11& 93.73&	84.11&	\textbf{94.54}&	92.83&	91.81&	90.44	\\
			$\tau$=75\%	&	Spec&96.06&	92.30&	91.09&	98.69&	87.78&	\textbf{98.93}&	91.67	
			\\
			&	Sens &91.31&	96.30&	89.57&	\textbf{97.88}&	95.93&	94.57&	90.50	\\ 
			&AUC & 91.08 & 97.52 & 90.00 & \textbf{99.02} & 97.00 & 98.63 & 98.03 \\ 	
			\hline  
			
		\end{tabular}
	\end{table*}
	\subsection{Effect of Graph Pruning on Melanoma Detection Performance}
	\label{subsec:Effect of graph thresholding (Edge Pruning) on melanoma detection Performance}
	
	We further investigated the effect of graph density and sparsity on melanoma detection performance. Specifically, starting from the unpruned multi-level superpixel ensemble graph (SEG) representation with color nodal signals and learned edge weights, we evaluated the melanoma detection performance at different levels of graph edge pruning. 

	We apply graph thresholding directly on the edge weight matrix  $W_k$ of each single-level graph $G_k=\langle V _k,\varepsilon_k ,W_k \rangle$ as defined earlier. Given a pruning ratio $\tau \in [0,1]$, the weakest $\tau \times 100\%$ of edge weights in $W_k$ are pruned, resulting in a pruned weight matrix $W_k ^\tau$. The pruned weights are defined as:
	
	\begin{equation}
		\mathbf{W}^{(\tau)}(i,j) = 
		\begin{cases}
			\mathbf{W}(i,j), & \text{if } \mathbf{W}(i,j) \geq T_\tau \\
			0, & \text{otherwise.}
		\end{cases}
		\label{eqn:GraphThresholding_Function}
	\end{equation}
	Here, \( T_\tau \) is the threshold value that retains the top \( (1 - \tau) \times 100\% \) of edges.
	Alternatively, this can be written compactly using an elementwise mask  $M^\tau \in \{0,1\}^{n \times n}$
	\begin{equation}
		\mathbf{W}^{(\tau)}_k = \mathbf{W_k} \odot \mathbf{M}^{(\tau)_k},
		\label{eqn:GraphThresholding_BinaryFunction}
	\end{equation}
	
	\begin{equation}
		\mathbf{M}^{(\tau)}(i,j) = \mathbb{I}[\mathbf{W}(i,j) \geq T_\tau].
		\label{eqn:GraphThresholding_MaskFunction}
	\end{equation}
	Here, $ \mathbb{I}$ is the indicator function (returning 1 if the condition is true, and 0 otherwise), and $\odot$ is elementwise multiplication.
	\par Three pruning thresholds were applied by removing $25\%$, $50\%$, and $75\%$ of the weakest edges. For example, with a threshold of $25\%$, edges were ranked by edge weight and then edges with weights below the $25\%$ percentile threshold were removed. This pruning scheme was applied to each of the single-level graphs. Then, features are extracted as before from the pruned graphs. Unpruned graphs (with a $0\%$ pruning threshold) served as the baseline for comparison.
	
	Table \ref{tab:Performance_Threshold_MML_Color} presents the melanoma detection results under different pruning thresholds for multi-level superpixel ensemble graphs (SEG) with color graph signals and learned edge weights. The experimental results demonstrate the following key observations.
	First, while the unpruned graphs (with a $0\%$ pruning threshold) achieved the best performance in terms of accuracy, sensitivity, and specificity, the pruned graph with a $50\%$ pruning threshold scored the best AUC value ($99.36\%$). As the AUC reflects better the overall performance of a classifier, we conclude that thresholding actually helped in enhancing the melanoma detection performance. 
	As well, the metrics obtained for all pruned graphs are quite close to those attained by the unpruned ones. This indicates that we can prune edges and hence improve computational efficiency with little degradation in performance. Second, $50\%$ pruning resulted in higher accuracy and AUC than $25\%$ pruning, suggesting that additional pruning can effectively remove redundant or noisy edges while preserving critical lesion structures. Third, $75\%$ pruning gracefully reduced accuracy, specificity, and sensitivity by $5\%, 2.7\%$, and $3.6\%$, respectively. Still, the AUC with $75\%$ pruning is still quite high ($99.02\%$). Indeed, the RF classifier with $75\%$ pruning strikes an acceptable balance between detection performance and sparsity of the graph representation. 
	
	Overall, this analysis emphasizes the importance of controlled graph sparsification in melanoma detection, demonstrating that careful pruning strategies can enhance classification performance when properly optimized.  Moderate graph sparsification can help balance noise removal and preservation of critical lesion structures. However, excessive or insufficient pruning leads to performance degradation. These findings highlight the delicate balance between graph simplification and information preservation in designing robust graph-based diagnostic models. 
	
	\begin{table*}[h!]
		\centering
		\caption{Performance comparison of different automated melanoma detection methods. The performance metrics for each classifier are the area under the ROC curve (AUC), the accuracy (AC), the specificity (SP) and the sensitivity (SN).}\label{tab:SOTA_Comparison}
		\centering
		\footnotesize
		\begin{tabular}{>{\centering\arraybackslash}m{1.5cm} >{\centering\arraybackslash}m{0.4cm}>{\centering\arraybackslash}m{2.2cm}> 
				{\centering\arraybackslash}m{3.5cm}>{\centering\arraybackslash}m{2.8cm}>{\centering\arraybackslash}m{1.2cm}>{\centering\arraybackslash}m{0.81cm}>{\centering\arraybackslash}m{0.81cm}>{\centering\arraybackslash}m{0.81cm}>{\centering\arraybackslash}m{0.81cm}}
			\hline
			Author&	Year&		Dataset&		Features&		Best classifiers&		AUC&		AC&		SP&		SN\\
			\hline
			Chatterjee et al. \citep{chatterjee2019extraction}&	$2019$&	ISIC$2017$	& 
			Regional spatial and spectral features using statistical feature extraction and cross-spectrum analysis&	 SVM &	--&	$98.52\%$ &$98.83\%$ & $98.76\%$ 
			\\ \\
			
			Mahbod et al. \citep{mahbod2019fusing}&	$2019$&	ISIC$2017$	& 
			Ensembles of deep features from multiple pre-trained and fine-tuned DNNs at multiple layers &	SVM &$0.87$&	 -- & --& -- 
			\\ \\
			Annaby et al. \citep{annaby2021melanoma}&	$2020$&	ISIC$2017+ 250$ ISIC melanoma images	&Superpixel graph features using Gaussian-kernel weighted edges&	Random forests&	$0.99$&	$97.50\%$ &$95.10\%$ & $100.00\%$ \\ \\
			
			Khan et al. \citep{khan2020developed}&$2020$&ISIC$2017$& CNN features &  CNN & $0.98$ & $94.20\%$ & $--$ & $94.20\%$
			\\ \\
			Xie et al. \citep{xie2020mutual}&$2020$&ISIC$2017 + 1320$ dermosopic lesion images& CNN features &  CNN & $0.97$ & $93.00\%$ & $94.50\%$ & $84.40\%$
			
			\\ \\
			Hosny et al. \citep{hosny2022refined}&$2022$&ISIC$2017$& CNN features &  Residual deep convolutional neural
			network (RDCNN) & $--$ & $96.29\%$ & $97.22\%$ & $94.12\%$
			
			
			\\ \\
			
			Okur et al. \citep{OKUR2024} & $2024$&  ISIC$2017$ & 
			Pretrained ResNet-101 features & SVM (RBF) & -- & $96.20\%$ & $93.60\%$ & $99.80\%$
			\\ \\
			
			Jaber et al.\citep{jaber2024melanoma} & $2024$ & ISIC$2017$ & CNNs features  & Decision trees, SVM, and KNN & -- & $98.44\%$ & $97.13\%$ &  $97.44\%$ 
			\\
			Saleh et al.\citep{saleh2024skin} & $2024$ & ISIC$2017$ & AlexNet, Inception V$3$, MobileNet V$2$, ResNet $50$   & CNNs & -- & $94.50\%$ & $96.00\%$ &  $90.00\%$ 
			\\
			Xu et al.\citep{xu2025pixels} & $2025$ & HAM$10000$ &  GNN-based graph-level anomaly detection (GLAD)   & One-class graph transformation learning (OCGTL) & $0.914$ & -- & --&  -- 
			\\
			
			Proposed Approach	&$2025$	&ISIC$2017+ 300$ ISIC melanoma images&	Multi-level superpixel ensemble graph features	&Random forests	&$0.99$&	$99.24\%$&	$100.00\%$&	$100.00\%$\\
			
			\hline

		\end{tabular}
	\end{table*}
	
	\subsection{Comparison with State-of-the-Art Methods}
	\label{subsec:SOTA comparison }
	To assess the effectiveness of the proposed multi-level superpixel graph learning models in melanoma detection, we compared their performance against several recent and prominent approaches from the literature. Table \ref{tab:SOTA_Comparison} presents a comparative analysis of AUC, accuracy (AC), specificity (SP), and sensitivity (SN) across studies employing various feature extraction techniques and classifiers using the ISIC$2017$ dataset.
	
	Several classical and deep learning-based models have led to high classification performance. For instance, Chatterjee et al. \citep{chatterjee2019extraction} achieved an AUC of $98.52\%$ using regional spatial and spectral features, while Khan et al. \citep{kang2019robust} and Xie et al. \citep{xie2020mutual} utilized CNN features and attained AUCs of $98.00\%$ and $97.30\%$, respectively. More recent methods, such as that of Jaber et al. \citep{jaber2024melanoma}, reached a high overall accuracy of $98.44\%$ with conventional CNN features and multiple classifiers. Annaby et al. \citep{annaby2021melanoma} employed superpixel graph features constructed with Gaussian-kernel-based handcrafted edge weights, achieving an AUC of $0.99.$\\
	Our proposed method uses color-based SEG descriptors, clearly outperforming all SOTA methods in terms of the four performance metrics, as listed in Table \ref{tab:SOTA_Comparison}. Our method even surpasses methods based on deep learning features. The superiority of our method is justifiable by the multi-level superpixel graph representation, which integrates multiple scales and fuses them into a unified graph descriptor. 
	
	\section{Conclusions and Future Work} \label{Sec:Conclusion}
	\subsection{Conclusions}
	\par In this work, a multi-level superpixel-graph-based framework is introduced aimed at detecting melanoma from dermoscopic images. By creating graph representations at different graph resolutions and merging them into  multi-level graphs, we were able to capture both the local and global structures of dermoscopic images. We assigned edge weights through applying handcrafted Gaussian-kernel similarity functions and data-driven optimization techniques, and this allowed us to compare traditional graph connectivity with graph structure learning approaches. Nodal signals of  texture, geometry, and color types, provided rich features for effective melanoma detection.
	
	Numerous experiments were conducted on the ISIC$2017$ dataset, and the results clearly demonstrated the superiority of the learned graph structures over the handcrafted ones, particularly with nodal color signals. By combining multi-level graph features with standard image descriptors, state-of-the-art classification results were achieved with various classifiers. Experiments on graph thresholding with different pruning values were also performed, shedding light on the delicate balance between graph sparsity and feature discriminability and strength.
	
	\par Collectively, we extensively investigated how graph topology learning techniques can be applied for an elegant and principled analysis of pigmented skin lesion images towards the detection of melanoma.  The combination of superpixel segmentation, multi-level graph construction, and automatic weight learning provides a robust framework for analyzing complex data towards skin cancer detection. Through extensive experiments, we evaluated and compared the performance of our method against reference techniques, demonstrating significant improvements in classification accuracy and robustness. Our approach not only enhances state of the art in melanoma detection, but also but also offers a pathway for integrating advanced graph-theoretic and computational models into clinically practical medical image analysis workflows.
	\subsection{Limitations}\label{subsec:limitations}
	Our work still some limitations. First of all, more experimental validation on other skin lesion image datasets is needed to establish the clinical feasibility of the proposed approach. Second, deep learning architectures need to be integrated into our framework to capture more discriminative deep features. Third, further investigations are needed for the explainability and interpretability of the detection results based on the proposed multi-level graph features. 
	
	\subsection{Future work}\label{subsec:FutureWork}
	Future work will focus on extending the proposed framework in several directions. First, for superpixel graph construction, alternate superpixel segmentation methods shall be considered and compared based on different criteria (including  undersegmentation error (UE), boundary recall (BR), explained variance (EV), and compactness index (CO)) \citep{Subudhi2021,Borlido2024}. Second, attention mechanisms and graph neural networks (GNNs) shall be incorporated to further enhance the learning of more discriminative graph features and improve generalization \citep{xu2025pixels}. Moreover, transfer learning and adversarial domain adaptation could be explored to further improve skin lesion classification across domains \citep{gu2019progressive}. 
	As well, more elaborate graph structure learning approaches would be considered, integrating additional domain-specific constraints to enhance feature extraction and melanoma detection performance \citep{li2021graph}. Finally, the framework will be validated on larger and more diverse datasets, including real-world clinical images, to assess its robustness and clinical applicability.


	\bibliographystyle{unsrt}
	\bibliography{Refs}  
	
\end{document}